\documentclass{article}
\PassOptionsToPackage{numbers, sort,compress}{natbib}
\usepackage[final]{neurips_2022}
\usepackage[utf8]{inputenc} 
\usepackage[T1]{fontenc}    
\usepackage{url}            
\usepackage{booktabs}       
\usepackage{amsfonts}       
\usepackage{nicefrac}       
\usepackage{microtype}      
\usepackage[table,dvipsnames]{xcolor}
\usepackage{xspace}
\usepackage[colorlinks,citecolor=NavyBlue]{hyperref}       
\newcommand{\supcon}{SC}

\newcommand{\ourbiased}{biased contrastive model}

\newcommand{\selecmix}{SelecMix}
\newcommand{\biasedsupcon}{GSC}

\newcommand{\biass}{biased features}
\newcommand{\rebuttal}[1]{{\textcolor{black}{#1\xspace}}}

\usepackage{amsthm}
\usepackage{amsmath, amssymb}
\usepackage{mathrsfs}
\usepackage{graphicx}
\usepackage{cleveref}
\usepackage{wrapfig}
\Crefname{figure}{Fig.}{Figs.}
\Crefname{section}{Sec.}{Sec.}
\Crefname{equation}{Eq.}{Eqs.}
\Crefname{proposition}{Prop.}{Props.}
\Crefname{prop}{Prop.}{Props.}
\Crefname{theorem}{Thm.}{Thms}
\Crefname{lemma}{Lemma}{Lemmas}
\Crefname{definition}{Def.}{Defs}
\Crefname{algorithm}{Alg.}{Algs}
\Crefname{remark}{Remark}{Remarks}
\Crefname{example}{Example}{Examples}
\Crefname{corollary}{Corollary}{Corollary}

\usepackage{commath}
\usepackage{bbm}
\usepackage{booktabs}
\usepackage{verbatim}
\usepackage{subfigure}
\usepackage{multirow}
\usepackage{makecell}
\usepackage{bbm}
\usepackage{mathtools}
\usepackage{lipsum}
\usepackage{adjustbox}
\usepackage{arydshln}

\usepackage{pifont}
\usepackage{xcolor}
\usepackage{pdfrender}
\usepackage{colortbl}
\definecolor{forestgreen}{rgb}{0.13, 0.55, 0.13}

\usepackage{algorithm}
\usepackage{algorithmic}
\usepackage{fancyhdr}

\newcommand{\stdv}[1]{\scriptsize$\pm$#1}

\newcommand{\todo}[1][]{\textcolor{red}{\bf [TODO]}}
\newcommand{\ie}{i.e.,}

\usepackage{color, colortbl}
\definecolor{Gray}{gray}{0.9}
\definecolor{lightgray}{rgb}{0.83, 0.83, 0.83}

\newcommand{\paragrapht}[1]{\noindent\textbf{#1}}

\usepackage{amsmath,amsfonts,bm}

\def\eqref#1{equation~\ref{#1}}

\def\1{\bm{1}}

\DeclareMathAlphabet{\mathsfit}{\encodingdefault}{\sfdefault}{m}{sl}
\SetMathAlphabet{\mathsfit}{bold}{\encodingdefault}{\sfdefault}{bx}{n}

\def\gB{{\mathcal{B}}}

\def\gD{{\mathcal{D}}}

\def\gL{{\mathcal{L}}}

\def\gN{{\mathcal{N}}}

\def\gP{{\mathcal{P}}}

\DeclareMathOperator*{\argmax}{argmax}
\DeclareMathOperator*{\argmin}{argmin}

\title{SelecMix: Debiased Learning by \\Contradicting-pair Sampling}

\author{
  Inwoo Hwang$^1$
  \quad
  Sangjun Lee$^1$
  \quad
  Yunhyeok Kwak$^1$
  \quad
  \textbf{Seong Joon Oh}$^3$
  \\
  \textbf{Damien Teney}$^4$
  \quad
  \textbf{Jin-Hwa Kim}\thanks{Corresponding authors.}$\:\:^{12}$
  \quad 
  \textbf{Byoung-Tak Zhang}$^{\ast 1}$
  \\
  $^1$AI Institute, Seoul National University
  \quad
  $^2$NAVER AI Lab
  \\
  $^3$University of Tübingen
  \quad
  $^4$Idiap Research Institute
}

\begin{document}

\maketitle

\begin{abstract}
Neural networks trained with ERM (empirical risk minimization) sometimes learn unintended decision rules, in particular when their training data is biased, i.e., when training labels are strongly correlated with undesirable features.
To prevent a network from learning such features, recent methods augment training data such that examples displaying spurious correlations (\ie~\textit{bias-aligned} examples) become a minority, whereas the other, \textit{bias-conflicting} examples become prevalent.
However, these approaches are sometimes difficult to train and scale to real-world data because they rely on generative models or disentangled representations.
We propose an alternative based on mixup, a popular augmentation that creates convex combinations of training examples. 
Our method, coined \selecmix{}, applies mixup to \emph{contradicting pairs} of examples, defined as showing either (i)~the same label but dissimilar biased features, or (ii)~different labels but similar biased features. 
Identifying such pairs requires comparing examples with respect to unknown biased features.
For this, we utilize an auxiliary contrastive model with the popular heuristic that biased features are learned preferentially during training.
Experiments on standard benchmarks demonstrate the effectiveness of the method, in particular when label noise complicates the identification of bias-conflicting examples.
\end{abstract}

\section{Introduction}
\label{sec:introduction}
The inductive biases contributing to the success of deep neural networks (DNNs) can sometimes limit their capabilities for out-of-distribution (OOD) generalization.
DNNs are prone to learn simple, linear predictive patterns from their training data, sometimes ignoring more complex but important ones~\cite{shah2020pitfalls,teney2021evading}.
It has been suggested that the simplest correlations in the data are often spurious~\cite{dagaev2021too}.
A DNN relying on such simple, spurious patterns will therefore display poor OOD generalization.
Spurious correlations in a dataset are often the result of a selection bias, and such datasets are therefore said to be \emph{biased}.
This paper is about debiased learning, also known as debiasing, i.e., methods that prevent a network from relying on spurious correlations when trained on a biased dataset.

Biased datasets typically contain a majority of so-called \textit{bias-aligned} examples and a minority of \textit{bias-conflicting} ones. In bias-aligned examples, ground truth labels are correlated with both robust and biased features.\footnote{A feature is biased if it displays a pattern that is statistically predictive of the labels over the dataset, though not necessarily on every example. For instance, a blue background may be present in most (but not all) images of birds. These images are said to be \textit{bias-aligned}.}
In bias-conflicting examples, labels are correlated only with robust features. 
Clearly, the issues of models trained on biased datasets stem from the prevalence of bias-aligned samples.
Various approaches for debiased learning encourage models to ignore biased features.
Since the identification of biased features from i.i.d. data is ill-defined, it requires additional assumptions, or supervision from heterogeneous (non-i.i.d.) training samples~\cite{scholkopf2021toward}.

In this work, we approach debiased learning with the assumption that biased features are ``easier to learn'' than robust ones, meaning that they are incorporated in the model earlier during training~\cite{shah2020pitfalls,scimeca2022which} \footnote{See~\cite{zhang2022rich} for a discussion of the disputed relevance of this assumption to real-world data.}.
Existing works based on this heuristic typically train two models: (i)~an auxiliary model that purposefully relies on biased features, and (ii)~the desired debiased model.
The auxiliary one guides the training of the debiased one~\cite{sanh2020learning,nam2020learning}. 
For example, \citet{nam2020learning} trains the auxiliary model with a \emph{generalized cross-entropy} (GCE) loss~\cite{zhang2018generalized} that strengthens its reliance on biased, easy-to-learn features. 
The training of the debiased model either augments the data with novel bias-conflicting examples~\cite{kim2021biaswap,lee2021learning} or upweights existing ones~\cite{nam2020learning,liu2021just}.
On the one hand, upweighting-based methods are simple but their debiasing capabilities are limited when bias-conflicting examples are scarce.
On the other hand, augmentation-based methods rely on carefully tuned generative models or disentangled representations that are difficult to apply to real-world data.

We propose a simple and effective method based on mixup~\cite{zhang2018mixup}, a popular data augmentation method that creates convex combinations of randomly-chosen pairs of examples and their labels.
Our method, coined \selecmix{}, is an application of mixup to selected \emph{contradicting pairs} of examples to generate new bias-conflicting examples.
We define contradicting pairs as having either (i)~the same ground truth label but dissimilar biased features, or (ii)~different labels but similar biased features.
To compare examples with respect to their biased features and thereby identify the contradicting pairs, we train an auxiliary contrastive model with a novel \emph{generalized supervised contrastive} (\biasedsupcon) loss that amplifies the reliance on easy-to-learn features.
As a result, feature clustering in the embedding space serves as a good indicator of the similarity of the biased features.
We train the auxiliary model and the desired debiased model simultaneously. The auxiliary model identifies contradicting pairs, while the debiased model is trained on data augmented by \selecmix{}.
Compared to past approaches, our method generates bias-conflicting examples without generative models or disentanglement, while implicitly upweighting existing ones since they are frequently selected for the mixup. 

We evaluate our method on standard benchmarks for debiasing. Experimental results suggest that \selecmix{} consistently outperforms prior methods, especially when bias-conflicting samples are scarce.
In addition, our method maintains its superior performance in the presence of label noise that complicates the identification of bias-conflicting examples.
Ablation studies show advantages of both (i) the selective mixup strategy compared to other mixup variants, and (ii) the \biasedsupcon{} loss, which strengthens the reliance on biased features and allows measuring the similarity of examples with respect to the biased features.

\section{Related work}
\label{sec:relatedwork}
\paragraph{Debiasing with known forms of bias or bias labels.}
Early works on debiasing assume some knowledge about the bias.
Some methods require that each training example is provided with a \emph{bias label}, i.e., a precise value for its biased features~\cite{hong2021unbiased,tartaglione2021end,kim2019learning,sagawa2019distributionally,li2019repair}.
Other methods use knowledge of the general form of the bias, such as color or texture in images.
This information is typically used to design custom architectures~\cite{wang2018learning,bahng2019rebias,cadene2019rubi}. 
For example, ReBias~\cite{bahng2019rebias} uses a BagNet~\cite{brendel2018approximating} as an auxiliary model because it focuses on texture, which is assumed to be the biased feature. The auxiliary model guides the training of a debiased model robust to unusual variations in texture.

\paragraph{Debiasing with the easy-to-learn heuristic.}
A number of recent works assume that biased features are learned more quickly than robust ones~\cite{lee2021learning,nam2020learning}.
A popular approach is to train an auxiliary model that intentionally relies primarily on the easy-to-learn biased features, e.g., with the GCE loss~\cite{zhang2018generalized}.
The auxiliary model guides the training of a debiased model that focuses on other, presumably non-biased features.
For example, LfF~\cite{nam2020learning} learns biased and debiased models simultaneously. Bias-conflicting training examples are identified based on the relative losses from the biased and debiased models, then upweighted for the training the debiased model.
BiaSwap~\cite{kim2021biaswap} trains an image translation model to generate new bias-conflicting training examples.
Building on LfF, DFA~\cite{lee2021learning} disentangles robust and biased features and swaps them randomly for augmentation.
Disentanglement is however an ill-posed problem in itself~\cite{locatello2019challenging} and a challenge with real-world data.
In contrast, our method augments bias-conflicting data by mixing existing examples, without generative models nor disentangled representations.

\section{Method}
\label{sec:method}
\newcommand{\bx}{{\boldsymbol{x}}}
\newcommand{\by}{{\boldsymbol{y}}}
\newcommand{\bp}{{\boldsymbol{p}}}
\newcommand{\btheta}{{\boldsymbol{\theta}}}

We first describe the use of mixup as a debiasing strategy, 
while assuming that each training example is (unrealistically) provided with bias labels (\Cref{sec:mixup_for_debiasing}).
Then, we move to a realistic scenario where bias labels are not available.
We introduce an auxiliary contrastive model that identifies the biased features, which are assumed to be "easier to learn" than robust ones. This allows comparing examples with respect to these biased features (\Cref{sec:bias_amplified_contrastive_learning}).
Finally, we describe the complete \selecmix{} method that combines the selective mixup with our auxiliary biased model (\Cref{sec:selective_mixup}).

\begin{figure}[t!]
\centering
\includegraphics[width=.9\textwidth]{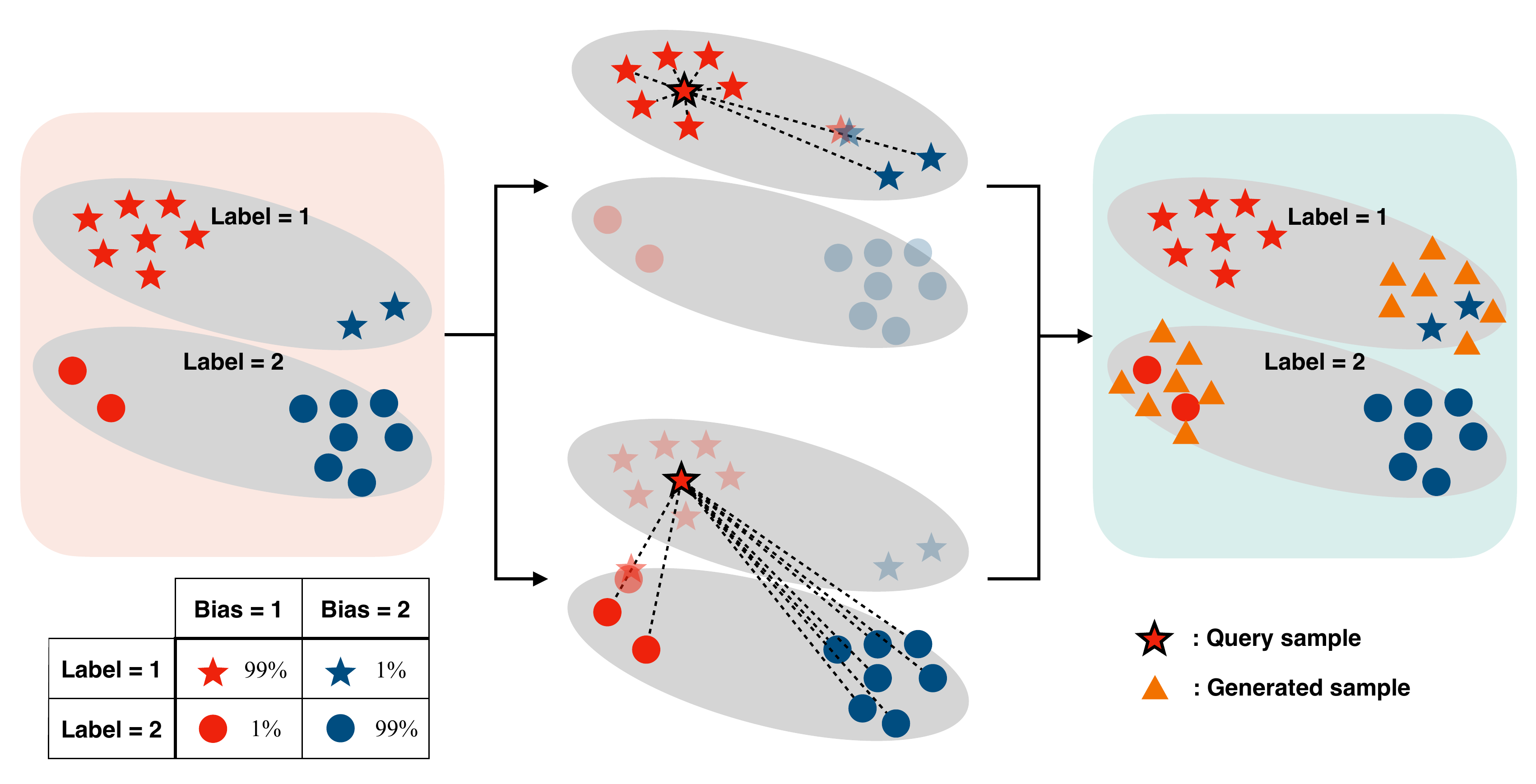}
\caption{Overview of the proposed SelecMix. A gray ellipse with the symbols represents a sharing label on the embedding space. The selective mixup is applied to the pairs of samples (dashed lines) having \textbf{(top)} the same label but dissimilar \biass, and \textbf{(below)} the different labels but similar \biass{} (Sec.~\ref{sec:selective_mixup}).
The orange triangles of the top and below ellipses represent the generated samples by the previous two procedures, respectively.
The legend illustrates the distribution of the exemplary dataset of two labels and two bias labels.
}
\label{fig:method_overview}
\end{figure}

\subsection{Mixup for augmenting bias-conflicting samples}
\label{sec:mixup_for_debiasing}
In a highly biased dataset, bias-conflicting samples constitute only a small fraction of the training data.
This is the root issue in the cases we consider.
The goal is thus to increase the fraction of bias-conflicting examples in the training data, which will then reduce the reliance of the learned model on biased features.
Our idea is to use mixup~\cite{zhang2018mixup} to augment the existing pool of bias-conflicting examples.
Mixup is a popular augmentation method~\cite{yun2019cutmix,kim2020puzzle,logitmix} known to improve various measures of robustness~\cite{zhang2020does,park2022unified}.
It constructs convex combinations of pairs of examples and their labels: 
\begin{equation}
\label{eq:Mixup}
(\widetilde{x}_i, \,\widetilde{y}_i) ~~\leftarrow~~ \big(\lambda \, x_i + (1\!-\!\lambda) \,x_{j}, ~~\lambda \, y_i + (1\!-\!\lambda) \, y_{j}\big),
\end{equation}
where $(x_i, y_i)$ and $(x_j, y_j)$ are two original training examples (e.g., image and one-hot label vector) and $\lambda$ is a random mixing coefficient, $\lambda\sim{U}\!\left[0,1\right]$.

Assuming for now that bias labels are available, we could generate bias-conflicting examples by applying mixup on pairs having either (i) the same ground truth label but different bias labels or (ii) different labels but the same bias label.
Any such pair includes at least one bias-conflicting example, so that mixup generates additional ones as long as this original example is assigned a higher mixing weight in \Cref{eq:Mixup}.
An overview of this mixup strategy is shown in \Cref{fig:method_overview}.
Next, we describe how to identify such desired pairs when bias labels are not available.

\begin{figure}[t!]
\centering
\includegraphics[width=\textwidth]{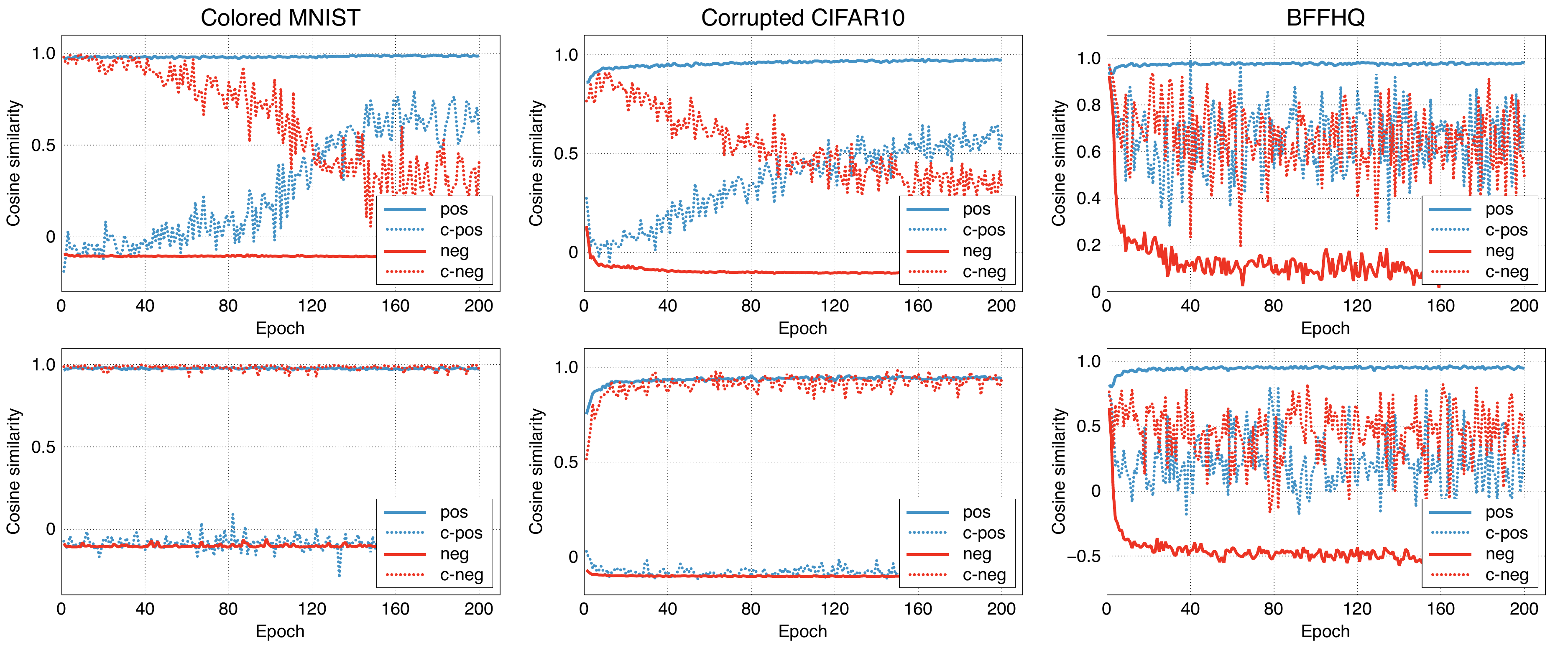}
\caption{Illustration of the similarity of positive, negative, and contradicting pairs when trained with \textbf{(top)} the \supcon~and \textbf{(bottom)} the proposed \biasedsupcon{} losses.
The solid line shows the average cosine similarity of 
(i) the pairs with the same label (\textit{positives}) and 
(ii) the pairs with different labels (\textit{negatives}). 
The dotted line represents (iii) the pairs with the same label but different bias labels (\textit{contradicting positives}) and (iv) the pairs with the different labels but the same bias label (\textit{contradicting negatives}). 
\textbf{Observations:}~As training proceeds with the \supcon~loss, the similarity of contradicting positives increases while it decreases for the contradicting negatives.
In contrast, the proposed \biasedsupcon{} loss amplifies the reliance on the biased features, thus the clustering in embedding space remains a good indicator of the similarity of the biased features during the whole training process.
}
\label{fig:sim_analysis_total}
\end{figure}

\subsection{Replacing bias labels with an auxiliary model}
\label{sec:bias_amplified_contrastive_learning}
We utilize an auxiliary model to compare training examples with respect to their biased features.
We assume that biased features are easier to learn than robust ones, because they are involved in simpler (e.g., linear) predictive patterns.
Our auxiliary model is trained to rely primarily on biased features.
We train the auxiliary model with a contrastive objective~\cite{he2020momentum,chen2020simple,khosla2020supervised}. This is known to induce a clustering in embedding space (better than standard cross-entropy) that reflects the similarity of training examples in terms of the learned features. 
These are \emph{biased} features by our assumption, such that the clustering reflects the similarity of examples w.r.t. their (unknown) bias labels.
We use the supervised contrastive (\supcon) loss of~\citet{khosla2020supervised}:
\begin{equation}
\label{eq:supcon}
\gL_{\textit{\supcon}} 
= -\sum_{i \in \gB} \frac{1}{|\gP_i|} \sum_{k \in \gP_i} \log p_{i, k}, \quad
\text{where}\quad p_{i, k} = \frac{\exp \left(\boldsymbol{z}_{i}^\intercal \boldsymbol{z}_{k} / \tau\right)}{\sum_{j \in \gB\backslash\{i\}} \exp \left(\boldsymbol{z}_{i}^\intercal \boldsymbol{z}_{j} / \tau\right)}, 
\end{equation}
where $\boldsymbol{z}_i$ is the normalized embedding of image $x_i$, 
$\gB=\{1, 2, ...,  B\}$ is the set of indices in the current mini-batch, 
$\gP_i=\{k\in\gB\backslash\{i\}\mid y_i=y_k\}$ is the set of positive examples relative to the example $i$ (i.e., with the same label), and the scalar $\tau$ is a temperature hyperparameter. 

As an experiment to confirm that the clustering of training examples in embedding space is indeed based on biased features, we train the auxiliary model on the Colored MNIST, the Corrupted CIFAR-10, and the BFFHQ datasets (see \Cref{sec:appendix_dataset_description}) with the \supcon~loss and compute the cosine similarity $\boldsymbol{z}_{i}^\intercal \boldsymbol{z}_{j}$ of the embeddings of all pairs of examples.
\Cref{fig:sim_analysis_total} shows that the examples are clustered according to the biased features early in the training. 
In other words, predictive patterns involving biased features are learned faster than those involving robust features, as desired. 
Since the higher cosine similarity $\boldsymbol{z}_{i}^\intercal \boldsymbol{z}_{j}$ implies a high probability $p_{i, j}$, we interpret it as the \emph{likelihood} of the pair $(i, j)$ having similar biased features.

To further amplify the reliance of the auxiliary model on the biased features, we define the \emph{generalized \supcon} (\biasedsupcon{}) loss as follows:
\begin{align}
\label{eq:biased_supcon}
\gL_{\textit{\biasedsupcon{}}}
&= - \sum_{i \in \gB} \, \frac{1}{|\gP_i|} \sum_{k \in \gP_i} \hat{p}_{i,k}^q \log p_{i, k}, 
\end{align}
where ${\hat{p}_{i,k}^q}$ is a scalar having the same value as $p_{i,k}^q$, meaning that the gradient is not back-propagated through it. 
The term ${\hat{p}_{i,k}^q}$ assigns a higher weight to sample pairs with a high probability ${{p}_{i,k}}$ and thus amplifies the reliance on the biased features.
We discuss the relationship between the GCE and \biasedsupcon{} losses in \Cref{sec:appendix_gce_gsupcon}.

\begin{figure}[t!]
\centering
\begin{minipage}{0.5\linewidth}
\begin{algorithm}[H]
\caption{\textsc{\selecmix{}}}
\label{alg:merge_selecmix}
\begin{algorithmic}[1]
\small
\STATE \textbf{Input}: batch $\gB=\{(x_i, y_i)\}_{i=1}^B$, biased model $g_\phi$
\STATE \textbf{Output}: batch $\widetilde{\gB}=\{(\tilde{x}_i, \tilde{y}_i)\}_{i=1}^B$
\STATE Sample $p\sim \text{U}[0,1]$, $\widetilde{\gB}=\varnothing$
\FOR{$i=1,\cdots,B$}
\STATE Sample $\lambda\sim \text{U}[0,1]$ and $\lambda \leftarrow \min(\lambda, 1-\lambda)$
\STATE if $p$ > $0.5$: $k = \underset{j\in \gP_i}{\argmin}\, g_\phi(x_i)^\intercal g_\phi(x_j)$
\STATE else: \quad $k = \underset{{j\in \gN_i}}{\argmax}\, g_\phi(x_i)^\intercal g_\phi(x_j)$ 
\STATE $(\widetilde{x}_i, \widetilde{y}_i) \leftarrow (\lambda x_i + (1-\lambda) x_{k}, y_k)$
\STATE $\widetilde{\gB} \leftarrow \widetilde{\gB} \cup \{(\widetilde{x}_i, \widetilde{y}_i)\}$
\ENDFOR
\end{algorithmic}
\end{algorithm}
\end{minipage}
\hfill%
\begin{minipage}{0.48\linewidth}
\begin{algorithm}[H]
\caption{Training with \textsc{\selecmix{}}}
\label{alg:merge_selecmix_full}
\begin{algorithmic}[1]
\small
\STATE \textbf{Input}: a dataset $\gD=\{(x_i, y_i)\}$, a model $f_\theta$, a biased model $g_\phi$, the number of iterations $T$
\STATE \textbf{Output}: a debiased model $f_\theta$
\STATE Initialize $\theta$ and $\phi$
\FOR{$t=1,\cdots,T$}
\STATE Draw a batch $\gB=\{(x_i, y_i)\}_{i=1}^B$ from $\gD$
\STATE $\widetilde{\gB} = \textsc{\selecmix{}}(\gB, g_\phi)$
\STATE Update $\theta$ with $\gL_{\text{CE}}(\widetilde{\gB})$
\STATE Update $\phi$ with $\gL_{\textit{\biasedsupcon{}}}(\gB)$
\hfill
\COMMENT{\Cref{eq:biased_supcon}} 
\ENDFOR
\end{algorithmic}
\end{algorithm}
\end{minipage}
\end{figure}

\subsection{Complete proposed method: selective mixup with biased embedding space}
\label{sec:selective_mixup}
We now have an auxiliary model for quantifying the similarity of the training examples in terms of the biased features.
We use it to apply mixup on \emph{contradicting pairs}, which have either (i)~the same label but dissimilar biased features (\textit{contradicting positives}) or (ii)~different labels but similar biased features (\textit{contradicting negatives}).

\paragrapht{Contradicting positives.} 
For each instance $(x_i, y_i)$ in the current mini-batch (i.e., the ``query''), we pick another one $(x_k, y_k)$ with the lowest similarity (measured in the space of their embeddings produced by the auxiliary model) among the set of positive examples (i.e., with the same label as $x_i$): 
\begin{equation}
\label{eq:contradicting_positive}
k = \underset{j\in \gP_i}{\argmin}\,\, p_{i, j} = \underset{j\in \gP_i}{\argmin}\,\, \boldsymbol{z}_{i}^\intercal \boldsymbol{z}_{j}, \quad \text{where} \,\, 
\gP_i = \{j\in\gB\setminus\{i\} \mid y_i=y_j\},
\end{equation}
where $\gB=\{1, 2, ...,  B\}$ is the set of the sample indices in the current mini-batch, and $\boldsymbol{z}_{i}$ and $\boldsymbol{z}_{j}$ are the normalized embeddings produced by our auxiliary model $g_\phi$, i.e., $\boldsymbol{z}_{i} = g_\phi(x_i)$.
Since we select the pair among the set of positives, the training CE loss of the mixed example is $l(\widetilde{x}_i, \widetilde{y}_i) = l(\lambda x_i + (1-\lambda) x_{k}, \lambda y_i + (1-\lambda) y_{k}) = l(\lambda x_i + (1-\lambda) x_{k}, y_k)$. 
Considering that most examples are bias-aligned in the training set, the query $(x_i, y_i)$ is also likely to be bias-aligned. 
In addition, since the query $x_i$ and the selected example $x_k$ have the same label but dissimilar biased features, it is also likely that the biased features of $x_k$ are not correlated with the label, i.e., the selected example $(x_i, y_i)$ is likely to be bias-conflicting. 
Thus, to effectively generate an example that contradicts the prediction based on biased features, we sample $\lambda\sim{U}\!\left[0,1\right]$ and assign the smaller value among $\lambda$ and $1-\lambda$ to $x_i$ and the larger one to $x_k$.

\paragrapht{Contradicting negatives.} For each query $(x_i, y_i)$, we pick another one with the highest similarity among the set of negative examples:
\begin{equation}
\label{eq:contradicting_negative}
k = \underset{{j\in \gN_i}}{\argmax}\,\, p_{i, j} = \underset{{j\in \gN_i}}{\argmax}\,\, \boldsymbol{z}_{i}^\intercal \boldsymbol{z}_{j}, \quad \text{where} \,\,
\gN_i = \{j\in\gB \mid y_i\neq y_j\}.
\end{equation}
Similarly, we sample $\lambda\sim{U}\!\left[0,1\right]$ and let $\lambda \leftarrow \min(\lambda, 1-\lambda)$.
In standard mixup, the training loss of the mixed sample is: $l(\widetilde{x}_i, \widetilde{y}_i) = l(\lambda x_i + (1-\lambda) x_{k}, \lambda y_i + (1-\lambda) y_{k}) = \lambda \cdot l(\lambda x_i + (1-\lambda) x_{k}, y_i) + (1-\lambda) \cdot l(\lambda x_i + (1-\lambda) x_{k}, y_k)$. 
However, considering that (i) the query $(x_i, y_i)$ is likely to be bias-aligned, and (ii) the pair $(x_i, x_k)$ shares similar \biass, the first term $\lambda \cdot l(\lambda x_i + (1-\lambda) x_{k}, y_i)$ acts as bias-aligned example since the biased feature shared with $x_i$ and $x_k$ is predictive of the label $y_i$.
Thus, rather than interpolating the label, we simply assign $\widetilde{y}_i \leftarrow y_k$.
The pseudo-code of the proposed algorithm is presented in \Cref{alg:merge_selecmix} and \Cref{alg:merge_selecmix_full}.

\paragraph{Intuitive interpretation.} 
Unlike standard mixup, we assign the label of the generated sample to the label of the selected sample, which is likely to be bias-conflicting.
Therefore, the proposed method can be viewed as \textbf{generating new bias-conflicting samples by injecting bias-aligned samples as noise to existing bias-conflicting samples}.
The method can also be interpreted as implicitly upweighting existing bias-conflicting samples, since they are frequently chosen for the mixup pair.

\section{Experiments}
\label{sec:experiment}
We now validate the effectiveness of the proposed method on debiasing benchmarks.
We also evaluate it under the presence of label noise, which is a challenging but realistic scenario, yet less explored so far (\Cref{sec:experiment_main_results}).
We also provide a detailed analysis of each component of our method:
(i)~the auxiliary contrastive model (\Cref{sec:experiment_biased_model}) and
(ii)~the selective mixup strategy (\Cref{sec:experimentforselecmix}).

\paragrapht{Datasets.}
The Colored MNIST is a modified MNIST~\cite{lecun-mnisthandwrittendigit-2010} that consists of colored images of ten digits where each digit is correlated with the color (e.g., most of the images of "0" are colored with red). Here, the label is a digit (i.e., $0\sim 9$) and the biased feature is color. 
The Corrupted CIFAR10 is constructed by applying different types of corruptions to the corresponding objects in the original CIFAR-10~\cite{krizhevsky2009learning} dataset (e.g., most of the images of dogs are corrupted with \textsc{Gaussian blur} noise).
The Biased FFHQ (BFFHQ)~\cite{lee2021learning} is constructed based on the real-world dataset FFHQ~\cite{karras2019style}, where the label is age and the biased feature is gender.
The ratio of bias-conflicting samples in the training set is $\alpha\in\{0.5\%, 1\%, 2\%, 5\%\}$ for \{Colored MNIST, Corrupted CIFAR10\} and $\alpha=0.5\%$ for BFFHQ.
All datasets are available in the official repository of DFA~\cite{lee2021learning}. 
We defer the detailed description of the datasets in \Cref{sec:appendix_dataset_description}.

\paragrapht{Baselines.} To evaluate the effectiveness of our method in debiasing, we compare it with the prior methods LfF~\cite{nam2020learning} and DFA~\cite{lee2021learning} which also rely on the easy-to-learn property of the biased features.
LfF trains an auxiliary model with the GCE loss to amplify its reliance on the biased features, then reweights examples for training a debiased model.
DFA disentangles the biased and robust features with a similar principle as LfF, then augments the data for training a debiased model by swapping the biased features across examples. 
We also include EnD~\cite{tartaglione2021end}, ReBias~\cite{bahng2019rebias}, and HEX~\cite{wang2018learning}.
EnD leverages explicit bias labels. ReBias and Hex are designed for a specific, known form of biased features such as color and texture in images.

\begin{table}[bht]
\caption{Main results. ($^\ast$)~Denotes methods tailored to predefined forms of bias, ($^\circ$)~methods using bias labels, and ($^\dagger$)~methods relying on the easy-to-learn heuristic. Numbers for HEX are from~\cite{lee2021learning}.}
\label{table:main_result}
\centering
\begin{adjustbox}{width=1.0\textwidth}
\begin{tabular}{cccccccccc}
\toprule
Dataset
& Ratio (\%)
& Vanilla
& HEX $^\textbf{*}$ \cite{wang2018learning}
& ReBias $^\textbf{*}$ \cite{bahng2019rebias}
& EnD $^\circ$ \cite{tartaglione2021end}
& LfF $^\dagger$ \cite{nam2020learning}
& DFA $^\dagger$ \cite{lee2021learning}
& V+Ours $^\dagger$
& L+Ours $^\dagger$
\\
\midrule
\multirow{4}{*}{\makecell{Colored \\MNIST}}
& 0.5 & {35.71}\stdv{0.83} & 30.33\stdv{0.76} & \textbf{71.42}\stdv{1.41} & {56.98}\stdv{4.85} & {63.86}\stdv{2.81} & {67.37}\stdv{1.61} & \underline{70.47}\stdv{1.66} & {70.00}\stdv{0.52} \\ 
& 1.0 & {50.51}\stdv{2.17} & 43.73\stdv{5.50} & \textbf{86.50}\stdv{0.97} & {73.83}\stdv{2.09} & {78.64}\stdv{1.51} & {80.20}\stdv{1.86} & \underline{83.55}\stdv{0.42} & {82.80}\stdv{0.71} \\ 
& 2.0 & {65.40}\stdv{1.63} & 56.85\stdv{2.58} & \textbf{92.95}\stdv{0.21} & {82.28}\stdv{1.08} & {84.95}\stdv{1.71} & {85.61}\stdv{0.76} & {87.03}\stdv{0.58} & \underline{87.16}\stdv{0.62} \\ 
& 5.0 & {82.12}\stdv{1.52} & 74.62\stdv{3.20} & \textbf{96.92}\stdv{0.09} & {89.26}\stdv{0.27} & {89.42}\stdv{0.65} & {89.86}\stdv{0.80} & {91.56}\stdv{0.17} & \underline{91.57}\stdv{0.20} \\ 
\midrule
\multirow{4}{*}{\makecell{Corrupted \\ CIFAR-10}} 
& 0.5 & {23.26}\stdv{0.29} & 13.87\stdv{0.06} & {22.13}\stdv{0.23} & {22.54}\stdv{0.65} & {29.36}\stdv{0.18} & {30.04}\stdv{0.66} & \underline{38.14}\stdv{0.15} & \textbf{39.44}\stdv{0.22} \\ 
& 1.0 & {26.10}\stdv{0.72} & 14.81\stdv{0.42} & {26.05}\stdv{0.10} & {26.20}\stdv{0.39} & {33.50}\stdv{0.52} & {33.80}\stdv{1.83} & \underline{41.87}\stdv{0.14} & \textbf{43.68}\stdv{0.51} \\ 
& 2.0 & {31.04}\stdv{0.44} & 15.20\stdv{0.54} & {32.00}\stdv{0.81} & {32.99}\stdv{0.33} & {40.65}\stdv{1.23} & {42.10}\stdv{1.04} & \underline{47.70}\stdv{1.35} & \textbf{49.70}\stdv{0.54} \\ 
& 5.0 & {41.98}\stdv{0.12} & 16.04\stdv{0.63} & {44.00}\stdv{0.66} & {44.90}\stdv{0.37} & {50.95}\stdv{0.40} & {49.23}\stdv{0.63} & \underline{54.00}\stdv{0.38} & \textbf{57.03}\stdv{0.48} \\ 
\midrule
\multirow{1}{*}{\makecell{BFFHQ}} 
& 0.5 & {56.20}\stdv{0.35} & 52.83\stdv{0.90} & {56.80}\stdv{1.56} & {56.53}\stdv{0.61} & {65.60}\stdv{1.40} & {61.60}\stdv{1.97} & \textbf{71.60}\stdv{1.91} & \underline{70.80}\stdv{2.95} \\ 
\bottomrule
\end{tabular}
\end{adjustbox}
\end{table}

\subsection{Main results} 
\label{sec:experiment_main_results}
We apply our method to a vanilla ResNet18 (V+Ours) and to LfF (L+Ours).
We use the Colored MNIST, Corrupted CIFAR10, and BFFHQ datasets. 
As shown in \Cref{table:main_result}, our method consistently outperforms baselines, except ReBias~\cite{bahng2019rebias} on the Colored MNIST. 
Their bias-capturing model BagNet~\cite{brendel2018approximating} is specifically tailored to color and texture as biased features by relying on local image patches as input.
Both DFA and our method augment the pool of bias-conflicting training examples, but DFA's reliance on disentangled representations seems problematic on the more complex datasets.
In contrast, our method performs well on all datasets, owing to the simplicity of the mixup strategy.
In particular, our method outperforms DFA by a large margin on BFFHQ while the gap is smaller on Colored MNIST where disentanglement is easier.
See \Cref{sec:appendix_experiment_setup} for experimental details.

\begin{figure*}[t!]
    \centering
    \subfigure[Colored MNIST]{
    \includegraphics[width=0.3\textwidth]{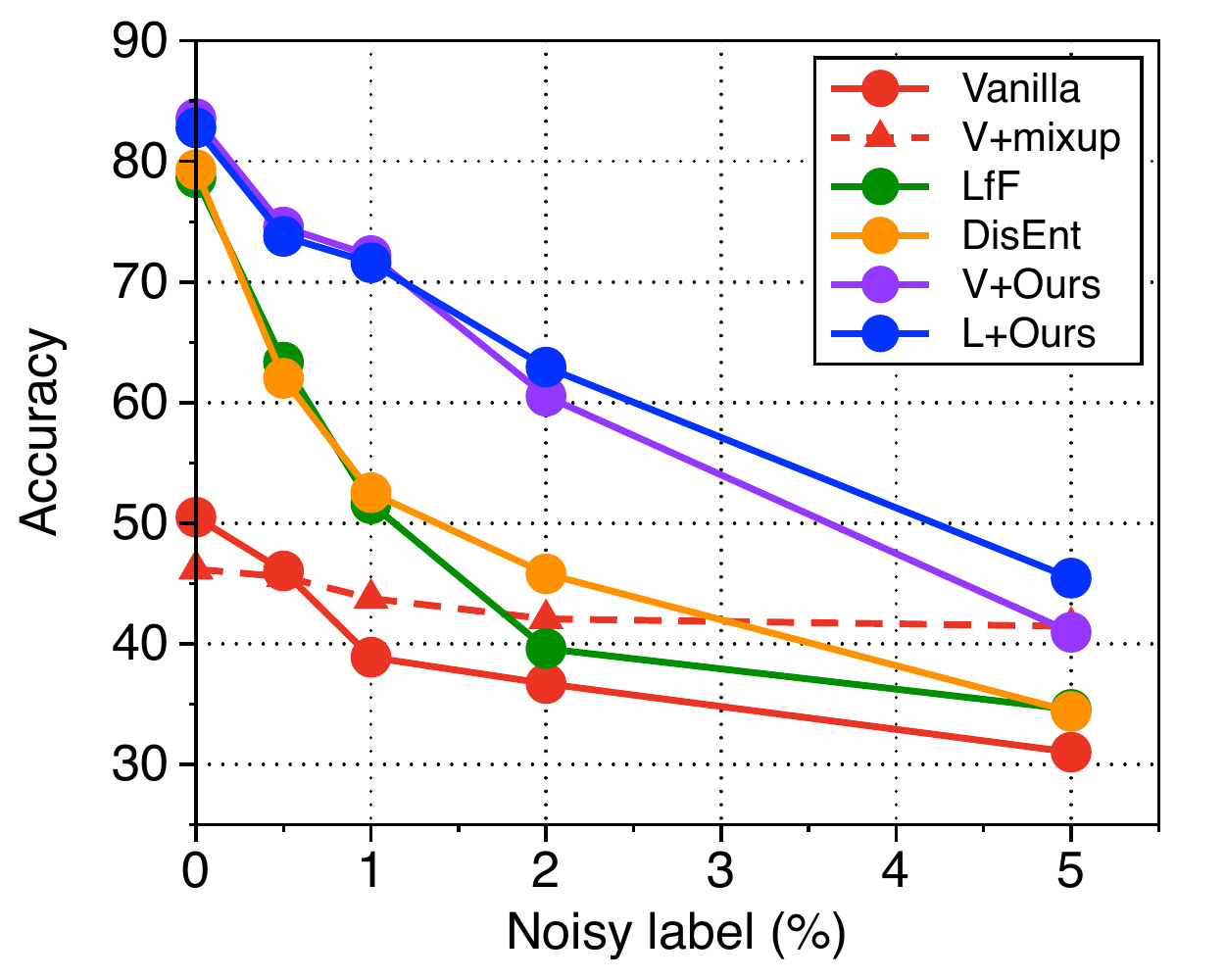}
    }
    \subfigure[Corrupted CIFAR-10]{
    \includegraphics[width=0.3\textwidth]{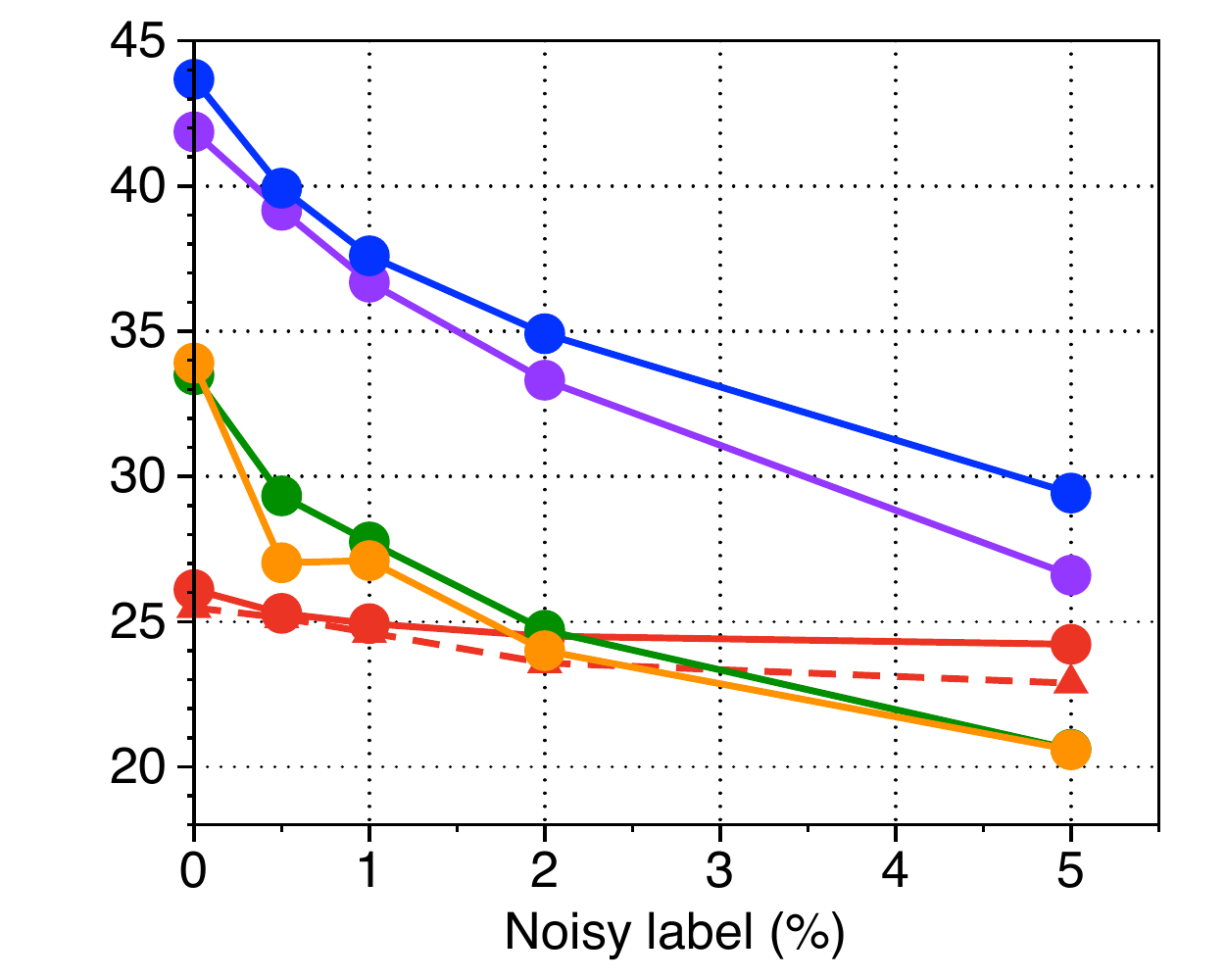}
    }
    \caption{Unbiased accuracy under the presence of label noise.}
    \label{fig:label_noise}
\end{figure*}

\paragrapht{Debiasing under the presence of label noise.}
\label{sec:experiment_label_noise}
Label noise can negatively affect debiasing methods that need to identify bias-conflicting examples. This is because training examples with noisy (i.e., incorrect) labels are difficult to distinguish from the bias-conflicting examples that we wish to identify. \Cref{fig:label_noise} shows that our method maintains better performance under the presence of label noise than baselines. We hypothesize that the robustness of our method comes from the nature of mixup, which is known to improve robustness~\cite{zhang2020does}.

\begin{table}[t]
\caption{Ablation study of various metrics on the biased embedding space. GT indicates that we used ground truth bias label for \selecmix{}. The cos and $l_2$ denote cosine distance and $l_2$ distance on the embedding space of the biased model trained with GCE loss, respectively. Ours denote the cosine distance on the embedding space of the proposed \ourbiased{}.}
\centering
\label{table:ablation_metric}
\begin{adjustbox}{width=0.87\textwidth}
\begin{tabular}{cccccc}
\toprule
\selecmix{} & GT & cos & $l_2$ & KL-divergence & Ours \\
\midrule
Contradicting positives & \underline{41.71}\stdv{0.85} & 40.31\stdv{0.63} & 38.56\stdv{1.25} & 40.82\stdv{0.94} & \textbf{42.94}\stdv{0.44} \\
Contradicting negatives & \underline{41.70}\stdv{0.74} & 33.69\stdv{0.97} & 34.15\stdv{0.50} & 30.99\stdv{0.93} & \textbf{41.82}\stdv{0.98} \\
Both & \underline{43.35}\stdv{0.67} & 38.05\stdv{0.83} & 38.90\stdv{0.96} & 39.94\stdv{1.48} & \textbf{43.68}\stdv{0.51} \\
\bottomrule
\end{tabular}
\end{adjustbox}
\end{table}

\subsection{Detailed analysis of the \ourbiased{}}
\label{sec:experiment_biased_model}

\paragrapht{Comparison of the biased embedding spaces for measuring bias similarity.}
\label{sec:experiment_ablation_metrics}
Our auxiliary contrastive model trained with the \biasedsupcon{} loss learns the embedding space which reflects the similarity of the examples w.r.t. their biased features. 
It is then used for \selecmix{} to identify the contradicting pairs by measuring the cosine similarity.
We replace our auxiliary model with the biased model of LfF~\cite{nam2020learning} which is trained with the GCE loss, and use it for \selecmix{}.
For the similarity measures, we used cosine distance and $l_2$ distance in the embedding space learned by their biased model, and the KL-divergence of the softmax outputs of the classification head.
As shown in \Cref{table:ablation_metric}, our auxiliary contrastive model achieves the best performance. Especially, the performance gap is significant for the contradicting negatives. 
This supports our claim in \Cref{sec:bias_amplified_contrastive_learning} that the proposed \biasedsupcon{} loss induces the feature clustering in the embedding space w.r.t. the biased features.
In contrast, the GCE loss learns the decision boundary and does not explicitly cluster the features, thus selecting the contradicting negatives, \ie~the pairs with different labels but similar \biass, seems to underperform ours.

\begin{table}[ht]
\caption{Bias label prediction accuracy.}
\label{table:bias_prediction}
\centering
\begin{adjustbox}{width=0.7\textwidth}
\begin{tabular}{lccccc}
\toprule
\multirow{2.5}{*}{Biased model} 
& \multicolumn{4}{c}{\makecell{Corrupted CIFAR-10}} 
& \multicolumn{1}{c}{\makecell{BFFHQ}} \\
\cmidrule(lr){2-5} \cmidrule(lr){6-6}
& 0.5\% & 1.0\% & 2.0\% & 5.0\% & 0.5\%
\\
\midrule
LfF \cite{nam2020learning} & 77.44\stdv{0.94} & 73.01\stdv{1.70} & 67.00\stdv{0.87} & 55.58\stdv{0.06} & 51.07\stdv{3.06} \\ 
DFA \cite{lee2021learning} & 79.03\stdv{1.15} & 72.30\stdv{0.71} & 64.71\stdv{0.24} & 52.01\stdv{0.70} & 46.20\stdv{1.00} \\ 
Ours & \textbf{95.45}\stdv{0.05} & \textbf{93.39}\stdv{0.02} & \textbf{92.89}\stdv{0.06} & \textbf{87.28}\stdv{0.20} & \textbf{59.13}\stdv{0.23} \\ 
\bottomrule
\end{tabular}
\end{adjustbox}
\end{table}

\begin{figure*}[t!]
    \centering
    \hspace{-.25in}
    \subfigure[LfF~\cite{nam2020learning}]{
    \includegraphics[width=0.255\textwidth, page=1]{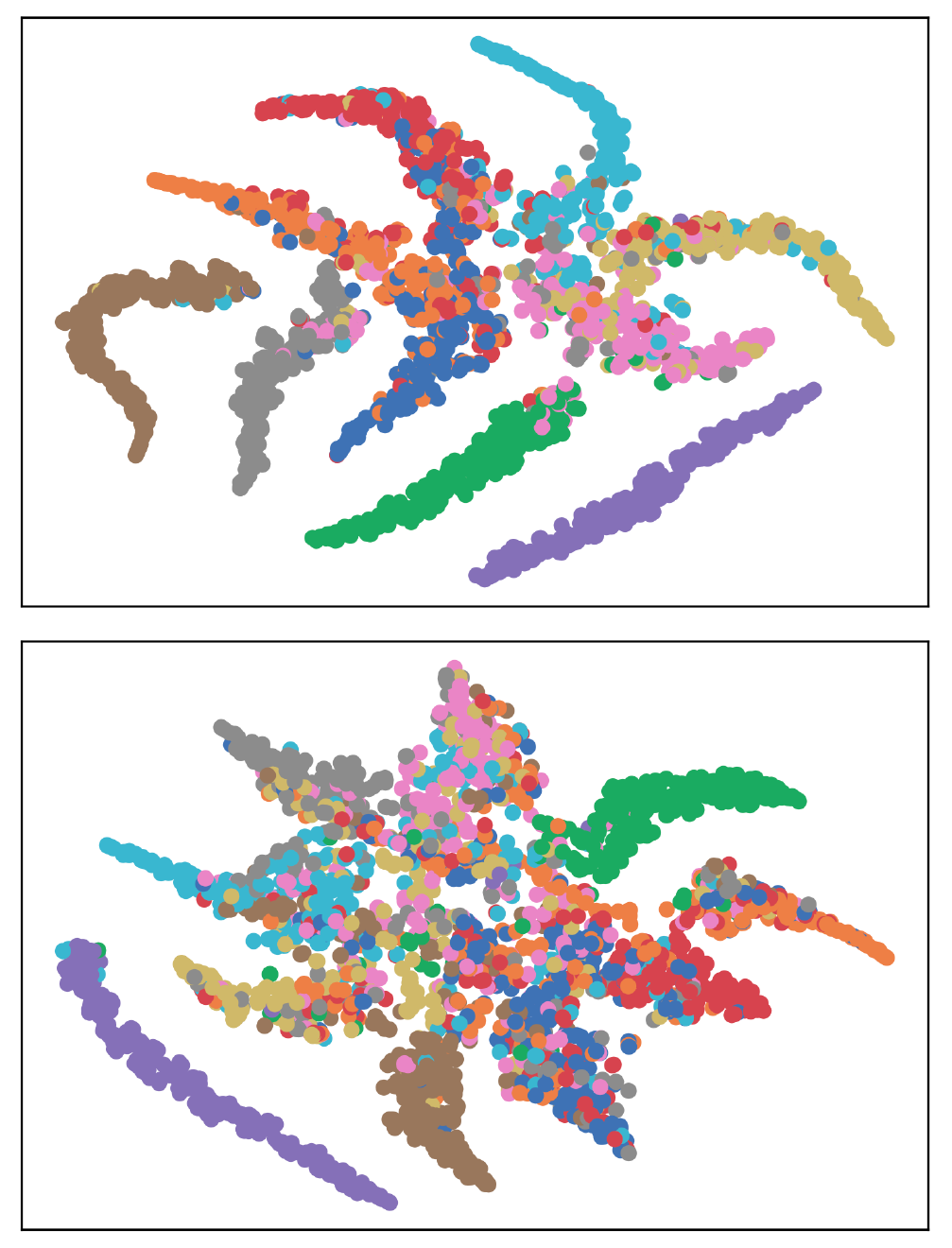}
    \label{fig:tsne_biased_a}
    }
    \hspace{-.2in}
    \subfigure[DFA~\cite{lee2021learning}]{
    \includegraphics[width=0.255\textwidth, page=2]{figure/tSNE/tsne_biased_all.pdf}
    \label{fig:tsne_biased_b}
    }
    \hspace{-.2in}
    \subfigure[\supcon~loss \cite{khosla2020supervised}]{
    \includegraphics[width=0.255\textwidth, page=3]{figure/tSNE/tsne_biased_all.pdf}
    \label{fig:tsne_biased_c}
    }
    \hspace{-.2in}
    \subfigure[\biasedsupcon{} loss]{
    \includegraphics[width=0.255\textwidth, page=4]{figure/tSNE/tsne_biased_all.pdf}
    \label{fig:tsne_biased_d}
    }
    \hspace{-.25in}
     \caption{Visualization with t-SNE of features extracted from (a)~the biased model of LfF \cite{nam2020learning}, (b)~the biased model of DFA \cite{lee2021learning}, (c)~the model trained with the \supcon~loss, and (d)~our auxiliary contrastive model trained with the \biasedsupcon{} loss. \textbf{(Top)} $\alpha$=$1\%$. \textbf{(Bottom)} $\alpha$=$5\%$.}
    \label{fig:tsne_biased}
\end{figure*}

\paragrapht{Bias label prediction of the biased model.}
\Cref{table:bias_prediction} shows the accuracy of the bias label prediction of the auxiliary biased model for each method. 
Our \ourbiased{} does not have a classification head since it is trained with a contrastive loss. Thus, we attach a linear classifier on top of the model and fine-tune it. 
Note that the bias labels are used only for the evaluation. 
As shown in \Cref{table:bias_prediction}, the \ourbiased{} shows the best performance.
As the ratio of the bias-conflicting samples increases, they degrade the performance. 
Notice our \ourbiased{} is robust to this effect compared with the prior methods relying on the GCE loss.

\paragrapht{Visualization.}
\Cref{fig:tsne_biased} illustrates t-SNE~\cite{van2008visualizing} plot of the features extracted from the corresponding models trained on the Corrupted CIFAR10. 
As shown in \Cref{fig:tsne_biased_d}, which corresponds to the proposed \ourbiased{}, we observe that the samples are well clustered on the embedding space of the \ourbiased{}.
Similar to the results in \Cref{table:bias_prediction}, the GCE-based biased models, \ie~LfF and DFA, suffer from the increasing number of bias-conflicting samples.

\begin{wrapfigure}[15]{R}{0.4\linewidth}
\centering
\includegraphics[width=\linewidth]{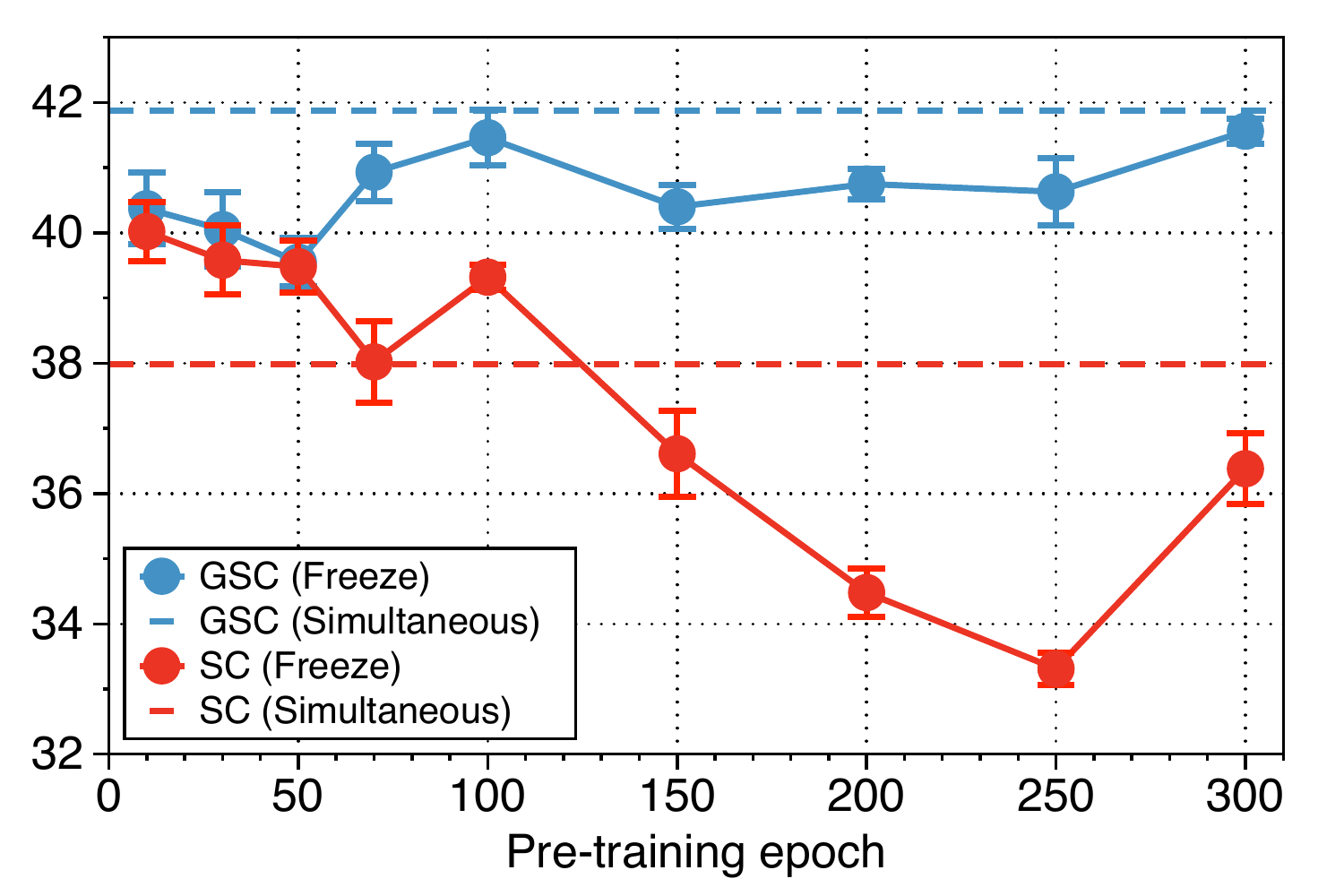}
\vspace{-10pt}
\caption{Comparison of simultaneous training vs. pretraining of the auxiliary model.}
\label{fig:freeze}
\end{wrapfigure}

\paragrapht{Pretraining vs. simultaneous training of the auxiliary model.}
For our method in the main experiments, the auxiliary biased model is simultaneously trained with the debiased model. 
To analyze the performance of the pretrained biased model, we equip \selecmix{} with the pretrained \ourbiased{} and evaluate the performance of our method with the varying total number of epochs for pretraining.
The solid line represents the unbiased accuracy of our method with the pretrained biased model. The dotted line represents the unbiased accuracy of our method, \ie~simultaneous training of the auxiliary biased model. 
As shown in \Cref{fig:freeze}, vanilla \supcon~loss exhibits the high variance of the performance with respect to the number of the epochs for pretraining, compared to the proposed \biasedsupcon{} loss.

\subsection{Detailed analysis of the selective mixup strategy of \selecmix{}}
\label{sec:experimentforselecmix}

\paragrapht{Ablation study.}
As shown in \Cref{table:ours_analysis}, we obtain the best performance when utilizing both the contradicting positives (A) and the contradicting negatives (B) all together in most cases.
This is because it generates more diverse bias-conflicting examples compared to when using only one of them alone.
On the other hand, standard mixup degrades the generalization capability in many cases which implies that the selection of the pairs is crucial. 

\begin{table}[t!]
\caption{Detailed analysis of \selecmix{}. (A) denotes \selecmix{} with only contradicting positives. (B) denotes \selecmix{} with only contradicting negatives. (AB) corresponds to the proposed \selecmix{}.}
\centering
\label{table:ours_analysis}
\begin{adjustbox}{width=0.99\textwidth}
\begin{tabular}{lccccccccccc}
\toprule
\multicolumn{1}{c}{\makecell{Dataset}}
& \multicolumn{4}{c}{\makecell{Colored MNIST}} & \multicolumn{4}{c}{\makecell{Corrupted CIFAR-10}} & \multicolumn{1}{c}{\makecell{BFFHQ}} \\
\cmidrule(lr){1-1} 
\cmidrule(lr){2-5} \cmidrule(lr){6-9} \cmidrule(lr){10-10}
\multicolumn{1}{c}{\makecell{Ratio (\%)}}
& 0.5
& 1.0
& 2.0
& 5.0
& 0.5
& 1.0
& 2.0
& 5.0
& 0.5
\\
\midrule
Vanilla & {35.71}\stdv{0.83} & {50.51}\stdv{2.17} & {65.40}\stdv{1.63} & {82.12}\stdv{1.52} & {23.26}\stdv{0.29} & {26.10}\stdv{0.72} & {31.04}\stdv{0.44} & {41.98}\stdv{0.12} & {56.20}\stdv{0.35} \\ 
+ mixup & {36.90}\stdv{1.87} & {46.19}\stdv{1.93} & {59.44}\stdv{3.61} & {76.28}\stdv{3.51} & {23.53}\stdv{0.80} & {25.48}\stdv{0.50} & {31.88}\stdv{0.42} & {43.41}\stdv{0.56} & {52.80}\stdv{0.40} \\ 
+ Ours (A) & {58.49}\stdv{0.77} & {74.59}\stdv{0.77} & {83.16}\stdv{1.14} & {89.32}\stdv{0.71} & \underline{37.16}\stdv{0.09} & \underline{41.38}\stdv{0.41} & \textbf{48.00}\stdv{0.64} & \textbf{54.40}\stdv{0.26} & {66.20}\stdv{1.80} \\ 
+ Ours (B) & \textbf{71.23}\stdv{1.30} & \underline{81.87}\stdv{0.94} & \underline{85.91}\stdv{0.51} & \underline{90.53}\stdv{0.81} & {35.17}\stdv{0.47} & {37.74}\stdv{0.17} & {42.77}\stdv{0.75} & {49.59}\stdv{0.05} & \underline{70.33}\stdv{0.46} \\ 
+ Ours (AB) & \underline{70.47}\stdv{1.66} & \textbf{83.55}\stdv{0.42} & \textbf{87.03}\stdv{0.58} & \textbf{91.56}\stdv{0.17} & \textbf{38.14}\stdv{0.15} & \textbf{41.87}\stdv{0.14} & \underline{47.70}\stdv{1.35} & \underline{54.00}\stdv{0.38} & \textbf{71.60}\stdv{1.91} \\ 
\midrule
LfF & {63.86}\stdv{2.81} & {78.64}\stdv{1.51} & {84.95}\stdv{1.71} & {89.42}\stdv{0.65} & {29.36}\stdv{0.18} & {33.50}\stdv{0.52} & {40.65}\stdv{1.23} & {50.95}\stdv{0.40} & {65.60}\stdv{1.40} \\ 
+ mixup & {44.30}\stdv{1.03} & {58.22}\stdv{2.71} & {72.44}\stdv{2.10} & {85.28}\stdv{1.39} & {22.71}\stdv{0.60} & {26.32}\stdv{1.10} & {32.67}\stdv{0.46} & {45.16}\stdv{0.92} & {57.53}\stdv{0.64} \\ 
+ Ours (A) & {57.28}\stdv{1.96} & {74.44}\stdv{0.36} & {84.20}\stdv{0.48} & {90.33}\stdv{0.22} & \underline{38.46}\stdv{0.40} & \underline{42.94}\stdv{0.44} & \underline{49.32}\stdv{0.28} & \underline{56.11}\stdv{0.83} & {67.60}\stdv{2.20} \\ 
+ Ours (B) & \textbf{70.26}\stdv{1.70} & \textbf{83.14}\stdv{0.86} & \underline{86.44}\stdv{0.49} & \underline{91.49}\stdv{0.31} & {37.15}\stdv{1.31} & {41.82}\stdv{0.98} & {47.01}\stdv{0.34} & {53.68}\stdv{0.02} & \underline{70.40}\stdv{2.23} \\ 
+ Ours (AB) & \underline{70.00}\stdv{0.52} & \underline{82.80}\stdv{0.71} & \textbf{87.16}\stdv{0.62} & \textbf{91.57}\stdv{0.20} & \textbf{39.44}\stdv{0.22} & \textbf{43.68}\stdv{0.51} & \textbf{49.70}\stdv{0.54} & \textbf{57.03}\stdv{0.48} & \textbf{70.80}\stdv{2.95} \\ 
\bottomrule
\end{tabular}
\end{adjustbox}
\end{table}

\begin{table}[t!]
\caption{Comparison with the mixup variants when the bias label is accessible.}
\centering
\label{table:ablation_mixup}
\begin{adjustbox}{width=0.99\textwidth}
\begin{tabular}{lccccccccccc}
\toprule
\multicolumn{1}{c}{\makecell{Dataset}}
& \multicolumn{4}{c}{\makecell{Colored MNIST}} & \multicolumn{4}{c}{\makecell{Corrupted CIFAR-10}} & \multicolumn{1}{c}{\makecell{BFFHQ}} \\
\cmidrule(lr){1-1} 
\cmidrule(lr){2-5} \cmidrule(lr){6-9} \cmidrule(lr){10-10}
\multicolumn{1}{c}{\makecell{Ratio (\%)}}
& 0.5
& 1.0
& 2.0
& 5.0
& 0.5
& 1.0
& 2.0
& 5.0
& 0.5
\\
\midrule
Vanilla & {35.71}\stdv{0.83} & {50.51}\stdv{2.17} & {65.40}\stdv{1.63} & {82.12}\stdv{1.52} & {23.26}\stdv{0.29} & {26.10}\stdv{0.72} & {31.04}\stdv{0.44} & {41.98}\stdv{0.12} & {56.20}\stdv{0.35} \\ 
\midrule
+ LISA (A) & {48.95}\stdv{0.75} & {67.10}\stdv{0.77} & {78.28}\stdv{0.99} & {87.04}\stdv{0.72} & {33.29}\stdv{0.65} & {38.62}\stdv{0.31} & {45.79}\stdv{0.13} & {53.41}\stdv{0.27} & {63.20}\stdv{0.92} \\ 
\rowcolor{Gray} 
+ Ours (A) & {58.07}\stdv{1.94} & {72.25}\stdv{0.60} & {82.35}\stdv{0.82} & {90.11}\stdv{0.14} & {36.10}\stdv{0.21} & {40.55}\stdv{0.45} & {46.70}\stdv{0.73} & {53.90}\stdv{0.59} & {67.67}\stdv{1.33} \\ 
\midrule
+ LISA (B) & {52.32}\stdv{2.30} & {72.97}\stdv{1.22} & {79.74}\stdv{2.25} & {86.44}\stdv{1.10} & {29.56}\stdv{0.93} & {34.23}\stdv{1.34} & {40.47}\stdv{0.67} & {47.61}\stdv{0.64} & {58.93}\stdv{0.50} \\ 
\rowcolor{Gray} 
+ Ours (B) & {72.02}\stdv{1.39} & {81.94}\stdv{1.62} & {86.04}\stdv{0.77} & {88.68}\stdv{2.32} & {35.47}\stdv{0.54} & {39.27}\stdv{0.87} & {43.13}\stdv{1.29} & {48.44}\stdv{0.96} & {77.40}\stdv{2.09} \\ 
\midrule
+ LISA (AB) & {60.85}\stdv{1.72} & {74.42}\stdv{2.27} & {83.20}\stdv{0.52} & {88.73}\stdv{0.39} & {32.71}\stdv{1.09} & {38.18}\stdv{0.90} & {44.15}\stdv{0.39} & {51.57}\stdv{0.45} & {65.20}\stdv{0.53} \\ 
\rowcolor{Gray} 
+ Ours (AB) & {70.57}\stdv{2.86} & {83.38}\stdv{0.53} & {87.22}\stdv{0.70} & {90.23}\stdv{0.58} & {37.02}\stdv{1.05} & {41.66}\stdv{1.10} & {48.35}\stdv{0.99} & {53.47}\stdv{0.53} & {75.00}\stdv{0.53} \\ 
\bottomrule
\end{tabular}
\end{adjustbox}
\end{table}

\paragrapht{Comparison with a mixup variant when bias label is available.}
We compare \selecmix{} with the recently proposed LISA~\cite{yao2022improving}, the mixup strategy for addressing domain generalization and subpopulation shift problems. 
LISA similarly applies the vanilla mixup on the pair of (i) the samples with the same label but different domains, and (ii) the samples with different labels but the same domain. Here, the domain label corresponds to the bias label in our experiment.
The key differences between \selecmix{} and LISA are (i) ours does not require the bias label, (ii) we do not interpolate the labels of the generated samples, and (iii) we sample $\lambda$ and let $\lambda \leftarrow \min(\lambda, 1-\lambda)$, \ie~we always assign the higher weight to the selected sample and the lower weight to the query sample (\Cref{sec:selective_mixup}).
\Cref{table:ablation_mixup} empirically confirms the better performance of \selecmix{}, using the explicit bias label for a fair comparison, over LISA, implying that augmenting the samples close to the bias-conflicting samples is crucial on the biased datasets. More discussions on the comparison between the \selecmix{} (using the explicit bias label) and LISA is provided in \Cref{sec:appendix_discussion_selecmix_lisa}.

\section{Conclusions}
\label{sec:conclusion}
We presented \selecmix{}, a method for debiased learning that augments bias-conflicting examples using mixup on contradicting pairs of examples. 
The selection of these pairs is the critical part of the method. 
It relies on an auxiliary model trained with a contrastive loss designed to amplify reliance on the biased features.
The biased features are assumed to be ``easy to learn'' and incorporated earlier than others during training by SGD.
\selecmix{} outperforms baselines on debiasing benchmarks and remains effective under the presence of label noise. 

\noindent\textbf{Limitations.}
The ``easy-to-learn'' assumption may not be correct, i.e., biased features are not always guaranteed to be learned faster than robust ones.
Our method should only be used when this assumption holds,
but it is unclear how to determine such cases a priori, and how frequent they are in real-world data~\cite{zhang2022rich}.
In the worst case, our method could {increase} reliance on biased features and \emph{worsen} robustness.
Our experiments use semi-synthetic datasets where the assumption is made valid by construction. Thus, our results are not a sign of the broad real-world applicability of the method.

\noindent\textbf{Societal impact.}
Our contributions should have a positive impact since our aim is to make machine learning systems more reliable. However, our method is only one step in this direction and the problems addressed should not be considered as solved.

\begin{ack}
We would like to thank Sangdoo Yun for the useful discussions and Yeonji Song for the suggestions on the writing. We also like to thank the anonymous reviewers for their constructive comments. 

This work was supported by the SNU-NAVER Hyperscale AI Center and the Institute of Information \& Communications Technology Planning \& Evaluation (2015-0-00310-SW.StarLab/10\%, 2019-0-01371-BabyMind/10\%, 2021-0-02068-AIHub/10\%, 2021-0-01343-GSAI/10\%, 2022-0-00951-LBA/10\%, 2022-0-00953-PICA/50\%) grant funded by the Korean government.
\end{ack}

\bibliographystyle{plainnat}
\bibliography{mainbib}

\section*{Checklist}

\begin{enumerate}

\item For all authors...
\begin{enumerate}
  \item Do the main claims made in the abstract and introduction accurately reflect the paper's contributions and scope?
    \answerYes{}
  \item Did you describe the limitations of your work?
    \answerYes{}
  \item Did you discuss any potential negative societal impacts of your work?
    \answerYes{}
  \item Have you read the ethics review guidelines and ensured that your paper conforms to them?
    \answerYes{}
\end{enumerate}

\item If you are including theoretical results...
\begin{enumerate}
  \item Did you state the full set of assumptions of all theoretical results?
    \answerNA{}
        \item Did you include complete proofs of all theoretical results?
    \answerNA{}
\end{enumerate}

\item If you ran experiments...
\begin{enumerate}
  \item Did you include the code, data, and instructions needed to reproduce the main experimental results (either in the supplemental material or as a URL)?
    \answerNo{}
  \item Did you specify all the training details (e.g., data splits, hyperparameters, how they were chosen)?
    \answerYes{See~\Cref{sec:appendix}}
        \item Did you report error bars (e.g., with respect to the random seed after running experiments multiple times)?
    \answerYes{}
        \item Did you include the total amount of compute and the type of resources used (e.g., type of GPUs, internal cluster, or cloud provider)?
    \answerYes{See~\Cref{sec:appendix}}
\end{enumerate}

\item If you are using existing assets (e.g., code, data, models) or curating/releasing new assets...
\begin{enumerate}
  \item If your work uses existing assets, did you cite the creators?
    \answerYes{}
  \item Did you mention the license of the assets?
    \answerYes{}
  \item Did you include any new assets either in the supplemental material or as a URL?
    \answerNo{}
  \item Did you discuss whether and how consent was obtained from people whose data you're using/curating?
    \answerNo{}
  \item Did you discuss whether the data you are using/curating contains personally identifiable information or offensive content?
    \answerNo{}
\end{enumerate}

\item If you used crowdsourcing or conducted research with human subjects...
\begin{enumerate}
  \item Did you include the full text of instructions given to participants and screenshots, if applicable?
    \answerNA{}
  \item Did you describe any potential participant risks, with links to Institutional Review Board (IRB) approvals, if applicable?
    \answerNA{}
  \item Did you include the estimated hourly wage paid to participants and the total amount spent on participant compensation?
    \answerNA{}
\end{enumerate}

\end{enumerate}


\newpage
\appendix

\section{Experimental details}
\label{sec:appendix}

\begin{wrapfigure}{R}{0.5\linewidth}
\centering
\vspace{-10pt}
\includegraphics[width=\linewidth]{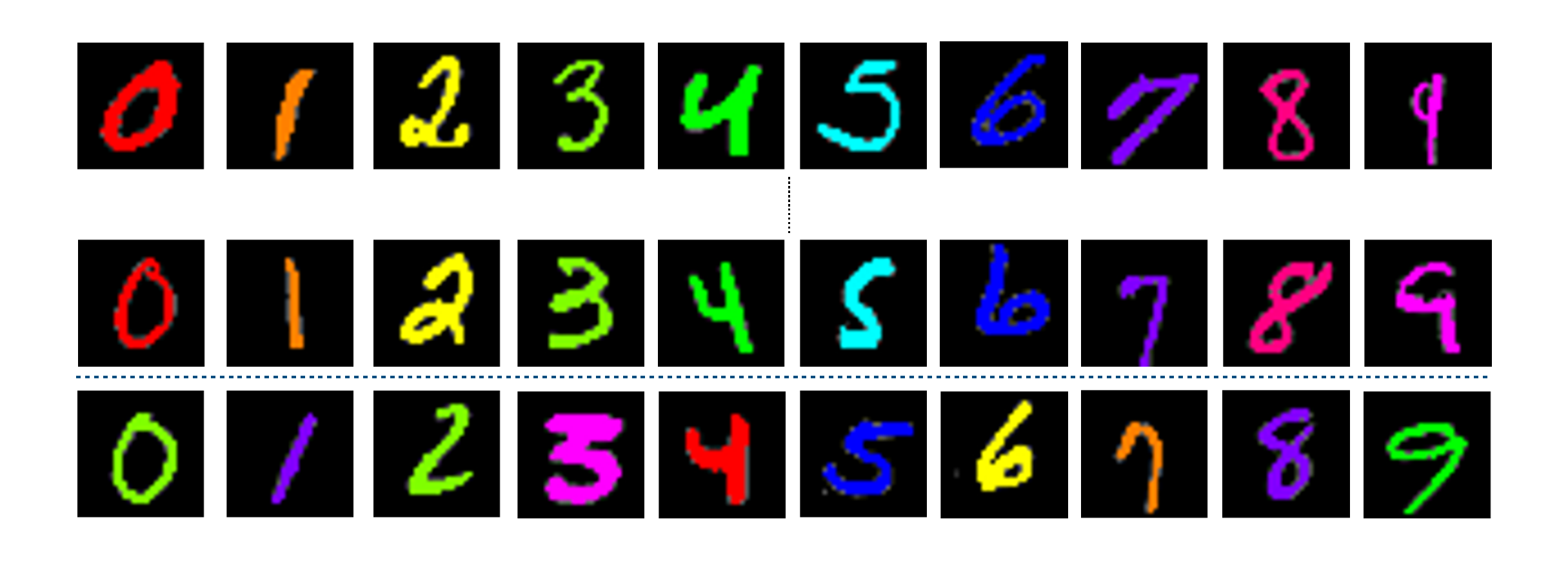}
\vspace{-15pt}
\caption{Example images of the Colored MNIST.}
\label{fig:appendix_dataset_cmnist}
\end{wrapfigure}

\subsection{A detailed description of datasets}
\label{sec:appendix_dataset_description}
\begin{figure*}[th]
\begin{minipage}{0.55\linewidth}
\includegraphics[width=\textwidth]{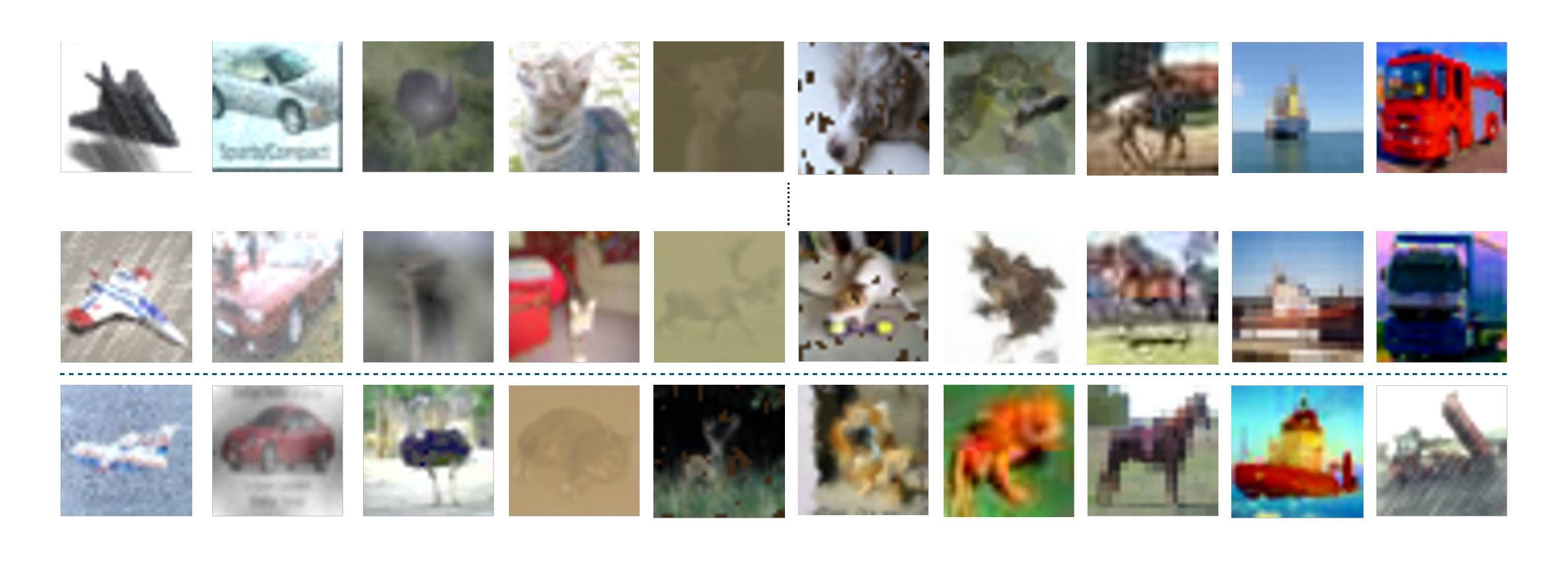}
\vspace{-10pt}
\caption{Example images of Corrupted CIFAR10.}
\label{fig:appendix_dataset_cifar10c}
\end{minipage}
\begin{minipage}{0.45\linewidth}
\includegraphics[width=\textwidth]{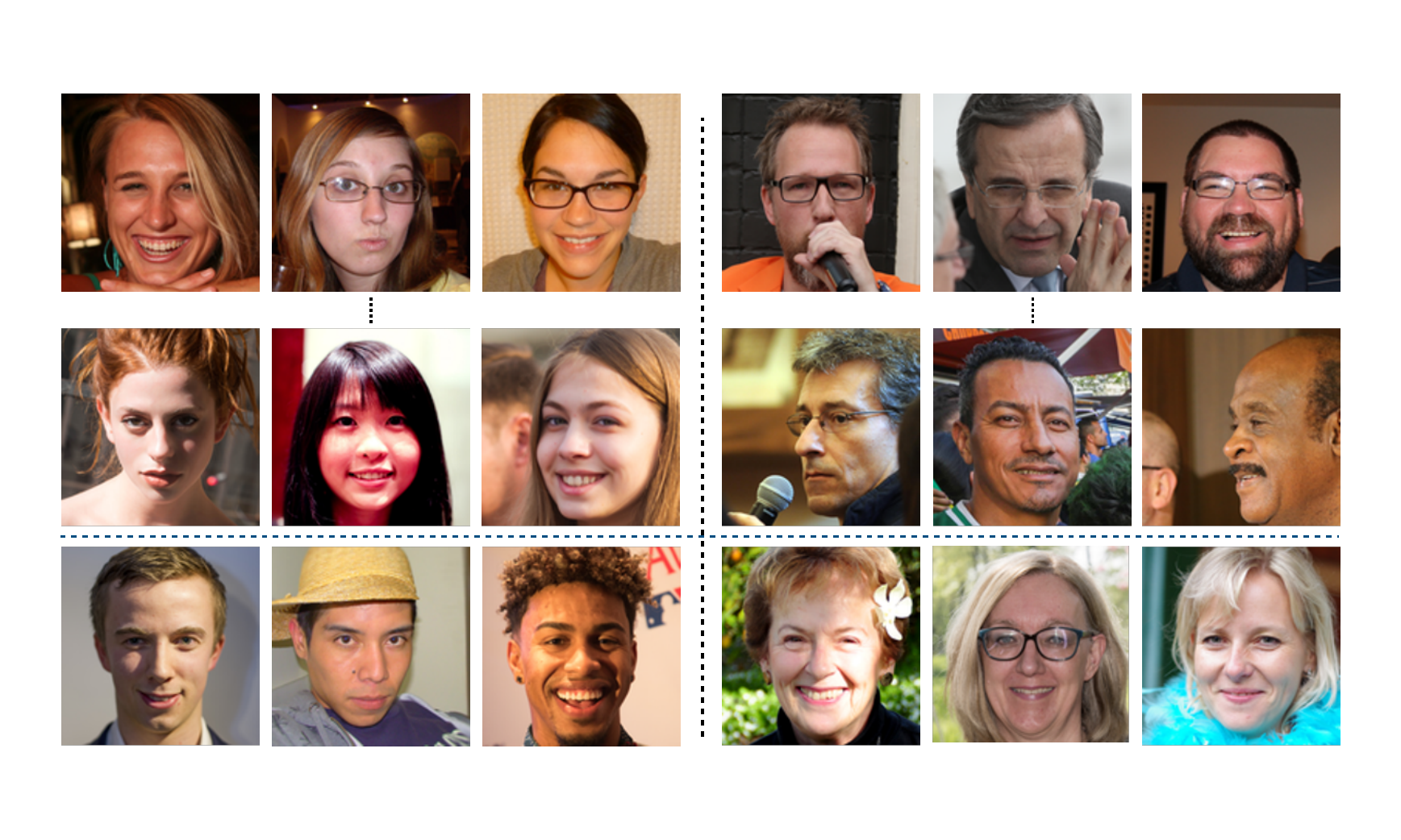}
\vspace{-20pt}
\caption{Example images of BFFHQ.}
\label{fig:appendix_dataset_bffhq}
\end{minipage}
\end{figure*}

For the main experiments, we evaluate our method on the Colored MNIST, Corrupted CIFAR-10, and BFFHQ. 
The example images of each datasets are shown in \Cref{fig:appendix_dataset_cmnist,fig:appendix_dataset_cifar10c,fig:appendix_dataset_bffhq}.
The Corrupted CIFAR-10 is a modified version of the CIFAR-10 dataset~\cite{krizhevsky2012imagenet} which is constructed by applying different types of corruptions. 
Specifically, each class is corrupted with one specific type of corruption from the following: \textsc{Brightness}, \textsc{Contrast}, \textsc{Defocus Blur}, \textsc{Elastic Transform}, \textsc{Frost}, \textsc{Gaussian Blur}, \textsc{Gaussian Noise}, \textsc{Impulse Noise}, \textsc{Pixelate}, and \textsc{Saturate}.
For the Colored MNIST and Corrupted CIFAR-10, we report the accuracy on the unbiased test set.
The BFFHQ~\cite{kim2021biaswap} is a subset of the FFHQ dataset~\cite{karras2019style} constructed for evaluating debiasing methods.
For the BFFHQ, we follow prior work~\cite{lee2021learning} and report the accuracy on the test set consisting of bias-conflicting examples.
The datasets are available in the official repository of DFA~\cite{lee2021learning}.

\subsection{Experimental details}
\label{sec:appendix_experiment_setup}
\paragraph{Setup.} 
We use a three-layer MLP for the Colored MNIST and a ResNet18~\cite{he2016deep} for \{Corrupted CIFAR-10, BFFHQ\} as the backbone architecture for all methods. 
We set the batch size to 256 and 64 for \{Colored MNIST, Corrupted CIFAR-10\} and \{BFFHQ\}, respectively. 
We train the models for 200 and 300 epochs for \{Colored MNIST, BFFHQ\} and \{Corrupted CIFAR-10\}, respectively.
To train our debiased model, we use the Adam optimizer with learning rates of 0.005, 0.001, and 0.0001 for the Colored MNIST, Corrupted CIFAR-10, and BFFHQ, respectively.
To train our auxiliary contrastive model, we use the Adam optimizer with learning rates of 0.01 and 0.001 for the Colored MNIST and Corrupted CIFAR-10, respectively, and use the SGD with a learning rate of 0.4 for the BFFHQ.
We did not use any additional augmentations to train the auxiliary model.
Most of the experiments were run on a single RTX 3090 GPU.

\paragraph{Implementation details.}
In the experiments, we combine our method with the vanilla ResNet18 (V+Ours) and LfF (L+Ours). 
To train the debiased model $f_\theta$, we update $\theta$ with $\gL_{\text{total}} = \lambda_{\text{base}}\gL_{\text{base}}(\gB) + \lambda_{\text{ours}} \gL_{\text{CE}}(\widetilde{\gB})$, where $\gB=\{(x_i, y_i)\}_{i=1}^B$ is the batch, $\widetilde{\gB}=\{(\tilde{x}_i, \tilde{y}_i)\}_{i=1}^B$ is the batch of the mixup samples (See \Cref{alg:merge_selecmix}), and $\gL_{\text{base}}$ is the update rule of the base algorithm for the debiased model (e.g., CE loss for the vanilla ResNet18).
We used $(\lambda_{\text{base}},\lambda_{\text{ours}})=(0, 1)$ for V+Ours, which corresponds to line 7 in \Cref{alg:merge_selecmix_full} (i.e., update $\theta$ with $\gL_{\text{CE}}(\widetilde{\gB})$) and  $(\lambda_{\text{base}},\lambda_{\text{ours}})=(1, 0.1)$ for L+Ours. 
Note that the training of the auxiliary biased model is agnostic of the base algorithm (e.g., Vanilla or LfF). 
We fixed the temperature $\tau=0.2$ (hyperparameter of the contrastive loss) and $q=0.7$ (hyperparameter of the GCE loss and \biasedsupcon{} loss) for all experiments.
For the major hyperparameters (i.e., $\lambda_{\text{base}}$, $\lambda_{\text{ours}}$, and $\tau$) of our method, we used the fixed values across the datasets and varying ratios.
For the hyperparameters of the baselines, we used their default choice, even if the optimal values may vary for the different ratios in the same dataset.
Since there is no provision for an unbiased validation set in most existing benchmarks, we follow the evaluation protocol of prior works~\cite{bahng2019rebias,nam2020learning,lee2021learning} and report the best accuracy (i.e., an ``oracle'' model selection).
Experimental results on the Corrupted CIFAR10 and the BFFHQ have averaged over 3 independent trials.
For the Colored MNIST, we report the average test accuracy over 5 independent trials.

\paragraph{Label noise experiment.}
\label{sec:appendix_label_noise_detail}
For the label noise experiments in \Cref{fig:label_noise}, we modify the Colored MNIST ($\alpha=1\%$) and the Corrupted CIFAR-10 ($\alpha=1\%$) by replacing the label with a random one (i.e., label noise) for a portion $\beta\in \{0.5\%, 1\%, 2\%, 5\%\}$ of the training examples. 
We used the same hyperparameters for the label noise experiments and the main experiments.

\section{Additional experiments}

\subsection{Identifying contradicting pairs} 
\begin{figure*}[t]
\centering
\subfigure[Contradicting positives]{
\includegraphics[width=0.35\textwidth]{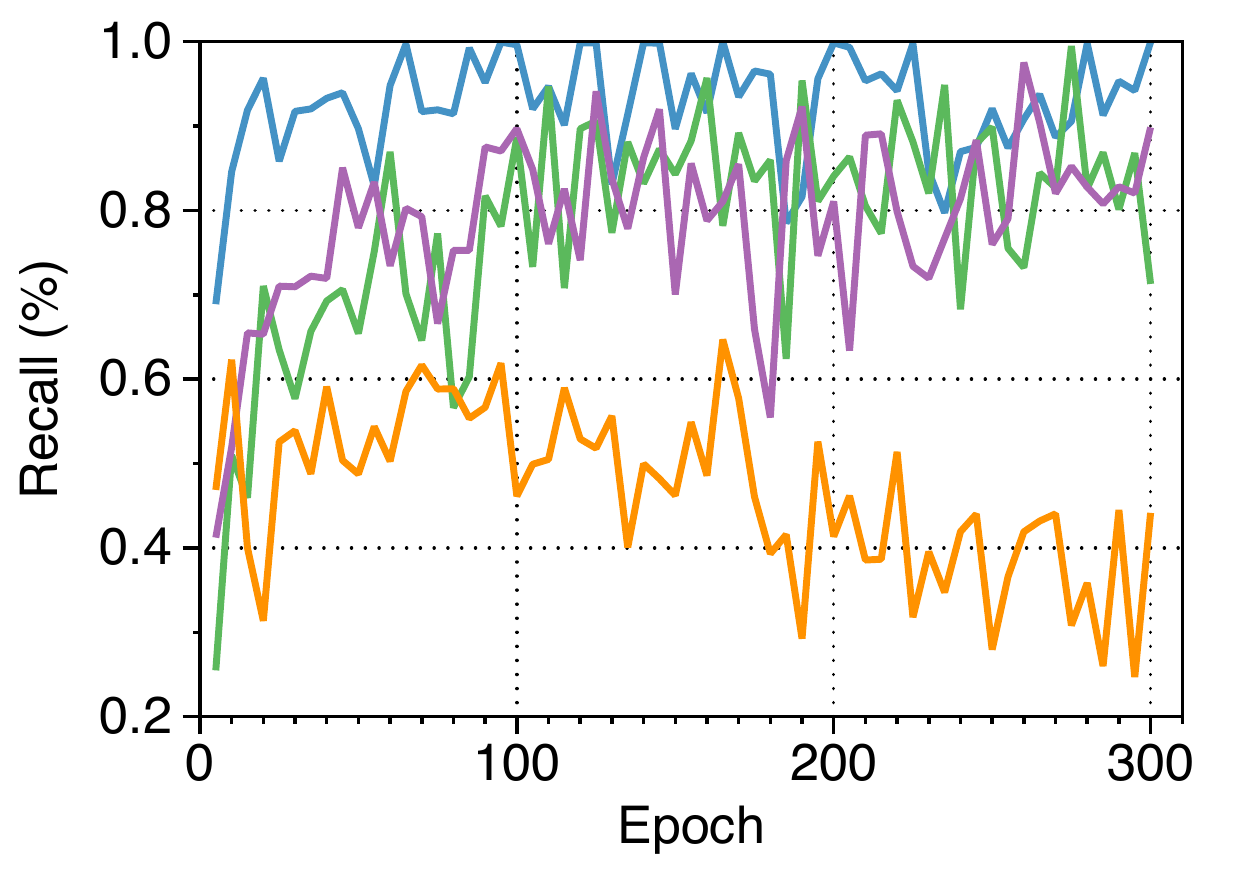}
}
\subfigure[Contradicting negatives]{
\includegraphics[width=0.35\textwidth]{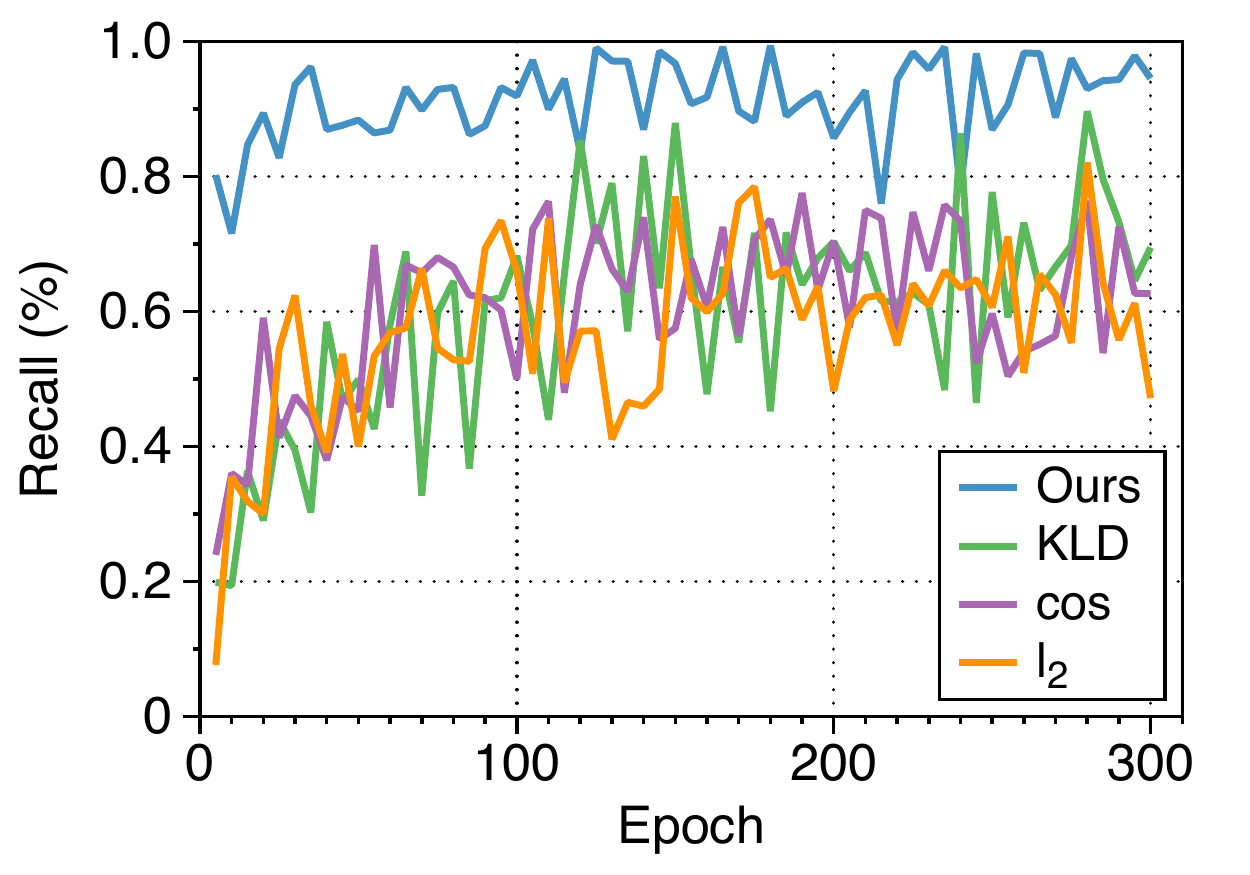}
}
\caption{Accuracy of discovering contradicting pairs. High recall implies that most of the contradicting pairs in the batch are discovered by \selecmix{}. The cos and $l_2$ denote respectively cosine and $l_2$ distance on the embedding space learned with the GCE loss.
KLD denotes the KL-divergence of the softmax outputs of the prediction head trained with the GCE loss.}
\label{fig:mixup-recall}
\end{figure*}

We compare our biased model with the model trained with the GCE loss, in terms of the accuracy of identifying contradicting positives and negatives on the Corrupted CIFAR-10 dataset.
As shown in \Cref{fig:mixup-recall}, our biased contrastive model better identifies the contradicting pairs, especially for discovering contradicting negatives. 
This implies that the \biasedsupcon{} loss effectively clusters the examples with respect to the biased features, which corroborates the results of \Cref{table:ablation_metric}.

\subsection{Bias-conflict accuracy}
In this subsection, we provide the bias-conflict accuracy (i.e., the accuracy of the bias-conflicting examples) of the baselines and our method. 
As shown in \Cref{table:bias_conflict_result}, our method outperforms the baselines in terms of the bias-conflict accuracy as well.

\begin{table}[h]
\caption{Bias-conflict accuracy of the baselines and our method.}
\label{table:bias_conflict_result}
\centering
\begin{adjustbox}{width=\textwidth}
\begin{tabular}{cccccccc} 
\toprule
\text{Dataset}
& \text{Vanilla}
& \text{Mixup~\cite{zhang2018mixup}}
& \text{EnD~\cite{tartaglione2021end}}
& \text{LfF~\cite{nam2020learning}}
& \text{DFA~\cite{lee2021learning}}
& \text{V+Ours}
& \text{L+Ours} \\
\midrule
\text{Colored MNIST} (1.0\%) & {44.14}\stdv{2.71} & {38.80}\stdv{3.85} & {62.92}\stdv{2.56} & {71.40}\stdv{2.63} & \underline{76.06}\stdv{3.15} & \textbf{77.38}\stdv{1.87} & {74.62}\stdv{0.94} \\ 
\text{Colored MNIST} (2.0\%) & {59.32}\stdv{1.52} & {51.88}\stdv{3.56} & {76.31}\stdv{1.16} & {80.82}\stdv{2.75} & {81.39}\stdv{1.97} & \textbf{83.32}\stdv{1.57} & \underline{82.70}\stdv{0.73} \\ 
\midrule
\text{Corrupted CIFAR10} (1.0\%) & {16.30}\stdv{0.50} & {14.77}\stdv{0.50} & {16.22}\stdv{0.53} & {24.08}\stdv{0.97} & {25.00}\stdv{2.55} & \underline{35.89}\stdv{1.05} & \textbf{36.17}\stdv{1.25} \\ 
\text{Corrupted CIFAR10} (2.0\%) & {21.40}\stdv{0.74} & {22.23}\stdv{0.89} & {23.77}\stdv{0.22} & {34.00}\stdv{1.06} & {31.84}\stdv{1.09} & \textbf{44.24}\stdv{1.07} & \underline{43.88}\stdv{1.41} \\ 
\bottomrule
\end{tabular}
\end{adjustbox}
\end{table}

\subsection{Comparison with JTT}
\rebuttal{In this subsection, we provide a comparison with JTT~\cite{liu2021just}. 
\Cref{table:jtt} shows the test accuracy of JTT and of our method. 
We used grid-search to choose the hyperparameters of JTT, i.e., the first stage training epoch $T$ and the upsampling ratio $\lambda$ as follows: $T\in\{1, 5, 10, 20, 30, 50\}$ and $\lambda\in\{5, 10, 30, 50, 100\}$.}

\begin{table}[t]
\caption{Comparison with JTT~\cite{liu2021just}.}
\label{table:jtt}
\centering
\begin{adjustbox}{width=0.6\textwidth}
\begin{tabular}{cccc}
\toprule
Dataset
& \text{Vanilla}
& JTT~\cite{liu2021just}
& \text{V+Ours}
\\ 
\midrule
\text{Colored MNIST} (1.0\%) & {50.51}\stdv{2.17} & {62.35}\stdv{3.30} & \textbf{83.55}\stdv{0.42} \\ 
\text{Colored MNIST} (2.0\%) & {65.40}\stdv{1.63} & {74.04}\stdv{1.33} & \textbf{87.03}\stdv{0.58} \\ 
\midrule
\text{Corrupted CIFAR10} (1.0\%) & {26.10}\stdv{0.72} & {28.55}\stdv{0.27} & \textbf{41.87}\stdv{0.14} \\ 
\text{Corrupted CIFAR10} (2.0\%) & {31.04}\stdv{0.44} & {33.03}\stdv{0.52} & \textbf{47.70}\stdv{1.35} \\ 
\midrule
\text{BFFHQ } & {56.20}\stdv{0.35} & {58.40}\stdv{0.35} & \textbf{71.60}\stdv{1.91} \\ 
\bottomrule
\end{tabular}
\end{adjustbox}
\end{table}

\subsection{Experiments on the UTKFace dataset}

\begin{wrapfigure}{R}{0.6\linewidth}
\centering
\vspace{-10pt}
\includegraphics[width=\linewidth]{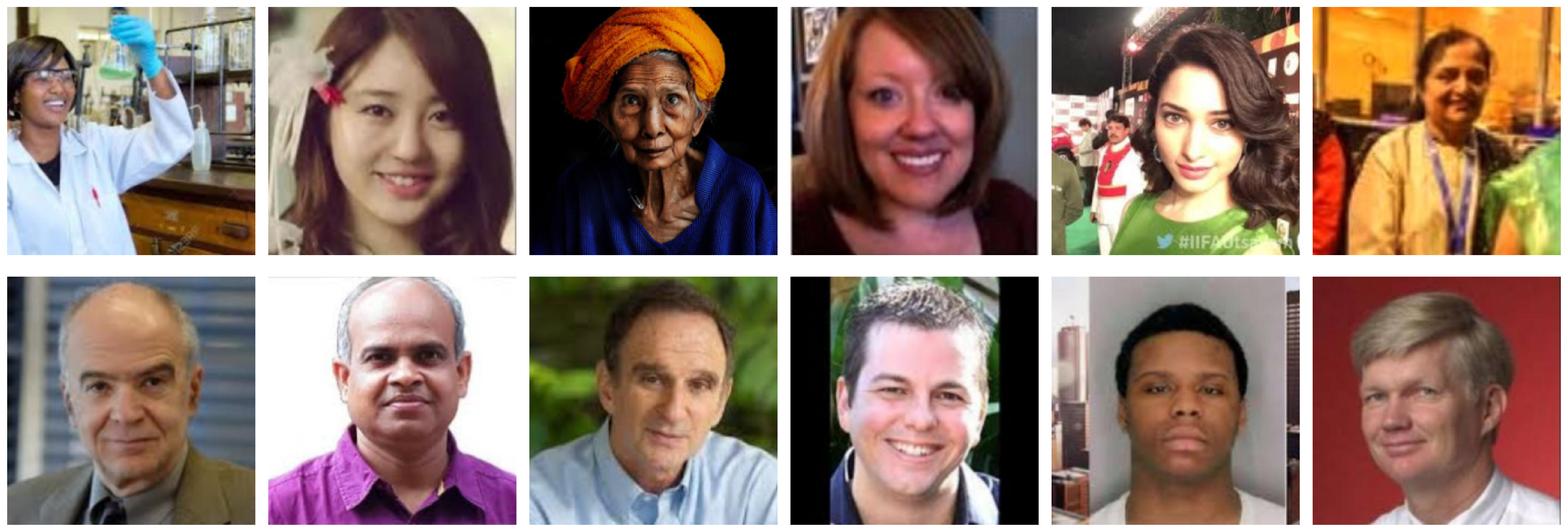}
\vspace{-15pt}
\caption{Example images of the UTKFace dataset.}
\vspace{-5pt}
\label{fig:appendix_utkface_examples}
\end{wrapfigure}

\paragraph{Dataset.} 
The UTKFace dataset is composed of various human face images. The dataset contains labels for race, age, and gender.
\Cref{fig:appendix_utkface_examples} shows the example images of the UTKFace dataset.
Following the prior work~\cite{hong2021unbiased}, we divide the samples into two groups for each label and compose a biased sub-dataset. We defer the details to~\cite{hong2021unbiased}.

\paragraph{Training details.} 
For the debiased model, we use a ResNet18 pretrained on the ImageNet dataset as a backbone architecture and use the Adam optimizer with a weight decay of 0.0001. 
We set an initial learning rate of 0.001 when the label is either gender or race, and 0.0005 when the label is age. 
We use the mixup of the contradicting positives for \selecmix{}. 
We use $(\lambda_{\text{base}},\lambda_{\text{ours}}) = (1,0.1)$. 
The learning rate is decayed by a factor of 10 at 1/3 and 2/3 of the total training epochs, following the prior work~\cite{hong2021unbiased}. 
For the auxiliary contrastive model, we use ResNet18 with the SGD optimizer.
We train the models for 40 epochs with a batch size of 128. 
We use the data augmentations of random resized crop and random horizontal flip.

\paragraph{Experimental results.}
As shown in \Cref{table:utk_result}, our method consistently outperforms the baselines for various compositions of target and bias features. This implies that the proposed method successfully scales to the complex real-world datasets which the prior works struggles on.

\begin{table}[h]
\caption{Unbiased accuracy of the baselines and our method trained on UTKFace dataset.}
\label{table:utk_result}
\centering
\begin{adjustbox}{width=0.6\textwidth}
\begin{tabular}{ccccc} 
\toprule
(Target, Bias) & Vanilla & LfF \cite{nam2020learning} & DFA \cite{lee2021learning} & V+Ours\\
\midrule
\multirow{1}{*}{\makecell{(Gender, Age)}} 
& \underline{74.09}\stdv{0.99} & 67.93\stdv{2.48} & 70.34\stdv{0.85} & \textbf{77.00}\stdv{0.66}\\
\multirow{1}{*}{\makecell{(Age, Gender)}} 
& \underline{92.00}\stdv{0.35} & 84.90\stdv{1.72} & 84.23\stdv{1.66} & \textbf{92.07}\stdv{0.10}\\

\multirow{1}{*}{\makecell{(Race, Gender)}} 
& \underline{77.73}\stdv{0.23} & 70.72\stdv{2.46} & 70.56\stdv{0.89} & \textbf{78.58}\stdv{0.49}\\
\multirow{1}{*}{\makecell{(Gender, Race)}} 
& \underline{88.68}\stdv{0.21} & 77.79\stdv{1.43} & 82.03\stdv{2.36} & \textbf{88.92}\stdv{0.11}\\

\bottomrule
\end{tabular}
\end{adjustbox}
\end{table}

\section{Additional discussions}

\subsection{Extended related work}
\label{sec:appendix_discussion_jm1}
JTT~\cite{liu2021just} proposed a 2-stage training framework to improve the group robustness without the group information. In the first stage, they train the vanilla network with ERM and construct the error set consists of the misclassified samples. In the second stage, they oversample examples of the error set to train the second network. 
JM1~\cite{giannone2021just} also constructs the error set and utilizes class-conditional mixup to further improve the group robustness.
In the transfer learning setting, \citet{bao2022learning} learns spurious features from multiple environments and constructs the metric space based on the spurious features. In contrast, our method exploits the easy-to-learn heuristic and learns the hypersphere embedding space with the proposed \biasedsupcon{} loss.

We now discuss the difference between JM1~\cite{giannone2021just} and our method, in particular, \selecmix{} (A).
While class-conditional mixup is utilized in both our method and JM1, the key difference is in the sampling of the pairs for the mixup. 
They combine the misclassified sample from the vanilla model with the sample that has the same label randomly chosen from the rest. 
On the other hand, SelecMix (A) combines the contradicting pair having the same label but \emph{the most dissimilar} biased features, by measuring the similarity with the biased model.

\subsection{Discussion on the relationship between GCE and \biasedsupcon{} losses}
\label{sec:appendix_gce_gsupcon}
In this subsection, we discuss the relationship between the generalized cross-entropy (GCE) loss~\cite{zhang2018generalized} and the proposed \biasedsupcon{} loss.
To begin with, we first explain how the GCE loss amplifies the reliance of the auxiliary model on biased features, compared to the standard cross-entropy (CE) loss. 

The GCE loss is defined as $\mathcal{L}_\textrm{GCE}(\bp,y)=(1-\bp_y^q) \,/\, q$ where $\bp$ is the softmaxed vector of predictions from the model and $\bp_y$ its $y^\textrm{th}$ component, $y$ is the ground truth class ID, and $q\in(0,1]$ a scalar hyperparameter.

The GCE loss simplifies to the CE loss as $q\!\rightarrow \!0$.
Assuming that the predictions are produced by a model of parameters $\btheta$, the gradients of the GCE and CE losses are related as follows:
\begin{equation}
\frac{\partial}{\partial \theta} \gL_{\textrm{GCE}_\theta}(\bp,y)
~~=~~ \bp_y^q \,.\, \frac{\partial}{\partial \theta} \gL_{\textrm{CE}_\theta}(\bp,y).
\end{equation}
Here, the term $\bp_y^q$ assigns the higher weight to the samples with a high probability $\bp_y$, thus upweights the "easy" samples and amplifies the reliance on the biased features.
While the model trained with the CE loss also focuses on the biased features since they are learned first, the GCE loss was shown to be more effective in identifying bias-conflicting examples by \citet{nam2020learning}.
Similar to the GCE/CE, our GSC loss improves over the SC loss~\cite{khosla2020supervised} in encouraging the model to rely primarily on biased features. The term ${\hat{p}_{i,k}^q}$ in~\Cref{eq:biased_supcon} plays the same role as the term $\bp_y^q$ in the GCE.

\subsection{A discussion comparing SelecMix with explicit bias labels and LISA }
\label{sec:appendix_discussion_selecmix_lisa}
As described in \Cref{sec:experimentforselecmix}, LISA~\cite{yao2022improving} is a similar mixup strategy that applies mixup on contradicting positives and negatives. 
When the explicit bias label is available, \selecmix{} also directly picks the contradicting pairs and applies mixup. Here, the key difference is that (i) \selecmix{} does not mix the labels, and (ii) \selecmix{} always assigns the higher weight to the selected sample. 
Thus, \selecmix{} more explicitly augment various bias-conflicting samples, compared to LISA.
As shown in \Cref{table:ablation_mixup}, the performance gap between \selecmix{} and LISA is significant for the small $\alpha$. This also implies that \selecmix{} is effective for debiased learning on the highly biased dataset.

\end{document}